\documentclass[sn-mathphys,Numbered]{sn-jnl}

\usepackage{graphicx}%
\usepackage{multirow}%
\usepackage{amsmath,amssymb,amsfonts}%
\usepackage{amsthm}%
\usepackage{mathrsfs}%
\usepackage[title]{appendix}%
\usepackage{xcolor}%
\usepackage{textcomp}%
\usepackage{manyfoot}%
\usepackage{booktabs}%
\usepackage{algorithm}%
\usepackage{algorithmicx}%
\usepackage{algpseudocode}%
\usepackage{listings}%

\usepackage{tabularx}
\usepackage{arydshln}
\usepackage{subfigure}
\usepackage[mathscr]{eucal}
\usepackage{diagbox} 
\usepackage{verbatim}
\usepackage{wrapfig} 
\usepackage{pifont}
\usepackage{bbm}
\usepackage{cleveref}

\definecolor{green(pigment)}{rgb}{0.1607, 0.3843, 0.0941}
\definecolor{blue(pigment)}{rgb}{0., 0.1484, 0.6992}
\usepackage{hyperref}

\newtheorem{thm}{Theorem}[section]
\newtheorem{mydef}[thm]{Definition}
\newtheorem{mycor}[thm]{Corollary}

\newtheorem{innercustomthm}{Proof}
\newenvironment{custompro}[1]
  {\renewcommand\theinnercustomthm{#1}\innercustomthm}
  {\endinnercustomthm}

\usepackage{caption}
\usepackage{varwidth}
\DeclareCaptionFormat{myformat}{%
  \begin{varwidth}{\linewidth}%
    \centering
    #1#2#3%
  \end{varwidth}%
}

\theoremstyle{thmstyleone}%
\theoremstyle{thmstyletwo}%

\theoremstyle{thmstylethree}%

\newcommand{\etr}{\mathcal{E}_{tr}}
\newcommand{\eall}{\mathcal{E}}

\newcommand{\St}{\mathcal{S}}
\newcommand{\D}{\mathcal{D}}
\newcommand{\R}{\mathcal{R}}

\newcommand{\argmin}{\mathrm{argmin}}

\newcommand{\rad}{\mathrm{Rad}}

\newcommand{\Var}{\mathrm{Var}}

\newcommand{\RR}{\mathbb{R}}
\newcommand{\EE}{\mathbb{E}}

\newcommand{\pr}{\mathbb{P}}
\newcommand{\x}{\mathbf{x}}
\newcommand{\y}{{\rm{y}}}

\newcommand{\xx}{\boldsymbol{x}}
\newcommand{\zz}{\boldsymbol{z}}
\newcommand{\yy}{y}
\newcommand{\w}{\boldsymbol{w}}
\newcommand{\vv}{\boldsymbol{v}}

\newcommand{\0}{\mathbf{0}}

\newcommand{\gaojie}[1]{{\color{red}{}#1}}

\begin{document}

\title[Article Title]{Invariant Correlation of Representation with Label: Enhancing Domain Generalization in Noisy Environments}


\author*[1]{\fnm{Gaojie} \sur{Jin}}\email{gaojie.jin.kim@gmail.com}

\author[2]{\fnm{Ronghui} \sur{Mu}}

\author[3]{\fnm{Xinping} \sur{Yi}}


\author[2]{\fnm{Xiaowei} \sur{Huang}}

\author[1]{\fnm{Lijun} \sur{Zhang}}

\affil[1]{\orgdiv{Key Laboratory of System Software (Chinese Academy of Sciences) and State Key Laboratory of Computer Science, Institute of Software}, \orgname{Chinese Academy of Sciences}, \orgaddress{\city{Beijing}, \country{China}}}

\affil[2]{\orgdiv{Department of Computer Science}, \orgname{University of Liverpool}, \orgaddress{\city{Liverpool}, \country{UK}}}

\affil[3]{\orgdiv{National Mobile Communications Research Laboratory}, \orgname{Southeast University}, \orgaddress{\city{Nanjing}, \country{China}}}


\abstract{
The Invariant Risk Minimization (IRM) approach aims to address the challenge of domain generalization by training a feature representation that remains invariant across multiple environments. 
However, in noisy environments, IRM-related techniques such as IRMv1 and VREx may be unable to achieve the optimal IRM solution, primarily due to erroneous optimization directions. 
To address this issue, we introduce ICorr (an abbreviation for \textbf{I}nvariant \textbf{Corr}elation), a novel approach designed to surmount the above challenge in noisy settings.
Additionally, we dig into a case study to analyze why previous methods may lose ground while ICorr can succeed. 
Through a theoretical lens, particularly from a causality perspective, we illustrate that the invariant correlation of representation with label is a necessary condition for the optimal invariant predictor in noisy environments, whereas the optimization motivations for other methods may not be.
Furthermore, we empirically demonstrate the effectiveness of ICorr by comparing it with other domain generalization methods on various noisy datasets.
The code is available at \url{https://github.com/Alexkael/ICorr}.
}




\maketitle

\section{Introduction}

Over the past decade, deep neural networks (DNNs) have made remarkable progress in a wide range of applications, such as computer vision~\cite{simonyan2014very,he2016deep,krizhevsky2017imagenet} and natural language processing \cite{bahdanau2014neural,luong2015effective}.
Typically, most deep learning models are trained using the Empirical Risk Minimization (ERM)~\cite{vapnik1991principles} approach, which assumes that training and testing samples are independently drawn from an identical distribution (I.I.D. assumption).
Nevertheless, recent studies have reported increasing instances of DNN failures~\cite{beery2018recognition,geirhos2020shortcut,degrave2021ai} when this I.I.D. assumption is violated due to distributional shifts in practice.

Invariant Risk Minimization \cite{DBLP:journals/corr/abs-1907-02893} is a novel learning approach that addresses the challenge of domain generalization (also known as out of distribution problem) in the face of distributional shifts.
The fundamental concept behind IRM is to train a feature representation that remains invariant across multiple environments~\cite{peters2016causal}, such that a single classifier can perform well in all of them.
Although obtaining the optimal invariant feature representation is challenging, previous works employ alternative methods~\cite{xu2021learning,shahtalebi2021sand,zhang2023free} to approximate it.
The success of IRM approach in existing training environments can ensure its ability to generalize well in new environments with unseen distributional shifts, which is evidenced by positive empirical results~\cite{rame2022fishr,chen2023pareto}.

\begin{figure*}[t]
\includegraphics[width=1.0
\textwidth]{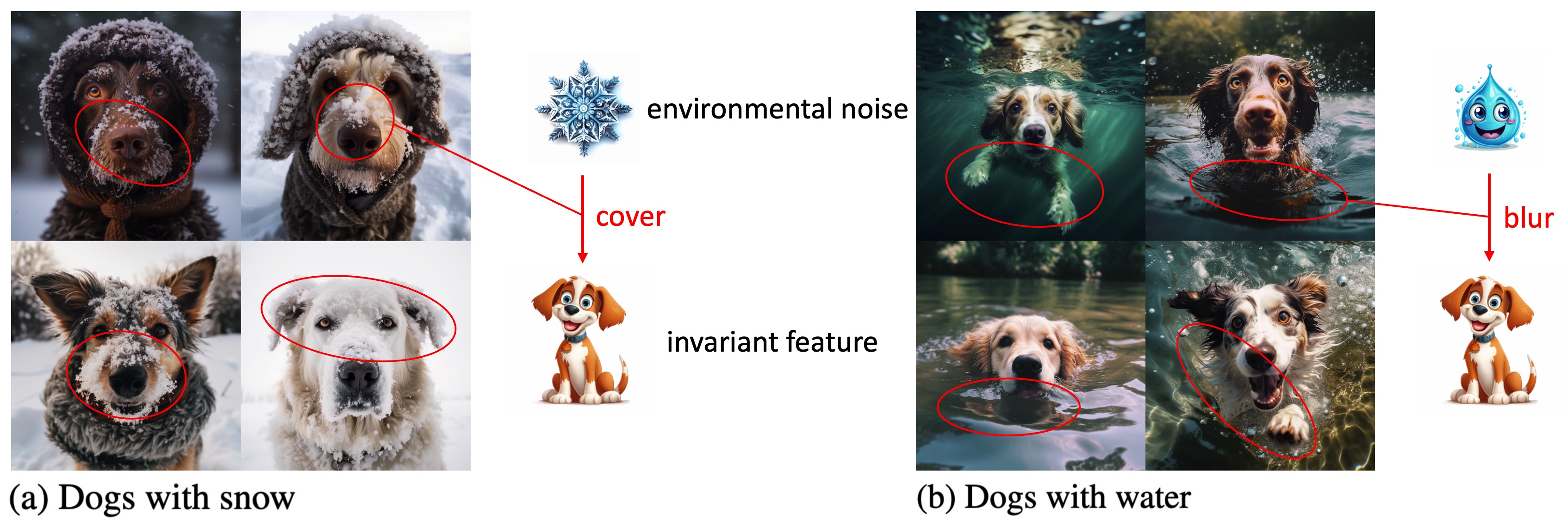}
\centering
\vspace{0mm}
\caption{A conceptual illustration of dogs in two environments: \textbf{(a)} snow, \textbf{(b)} water.
As snow may cover the hair of dogs and water may wet the appearance, they can cause different environmental inherent losses.
All images are generated by authors using Midjourney~(\url{www.midjourney.com}).}
\label{fig: example}
\vspace{-2mm}
\end{figure*}

However, in the real world, different environments (or domains) may exhibit varying levels of inherent (independent) noises, leading to various inherent losses.
Even an optimal IRM model cannot mitigate these inherent losses, resulting in varying optimal losses across different environments.
As shown in \Cref{fig: example}, inherent noise (such as snow or water) can impact the invariant feature (dog), such as covering the face or blurring the body, resulting in different inherent losses.
Existing IRM-related methods, such as IRMv1, VREx~\cite{krueger2021out} and \citet{rame2022fishr}, 
focus on optimizing the model in different clean environments but may fail in these noisy situations.

We conduct an analysis in this study to identify the reasons why existing IRM-related methods may be ineffective in noisy environments. 
Upon examining the case study presented in \Cref{sec: case study}, it has come to our attention that the optimization methods utilized for IRMv1, VREx and others 
may fail to converge to the optimal IRM solution due to environmental noise interference. 
Fortunately, our proposed method (ICorr) in \Cref{sec:ICorr} can effectively overcome these challenges,
because independent environmental noise should have no effect on the correlation between invariant representation and label. 
Following the theoretical setting from \citet{DBLP:journals/corr/abs-1907-02893} and \citet{peters2016causal}, we also provide in \Cref{sec:theory} a formal theoretical analysis from the perspective of causality, demonstrating that the invariant correlation across environments (i.e., the optimization idea of ICorr) is a necessary condition for the (optimal) invariant predictor in noisy environments, while the optimization motivations for others may not be.
Furthermore, in \Cref{sec: experiments}, we conduct a comprehensive range of experiments to confirm the effectiveness of ICorr in noisy environments.

We summarize the contributions and novelties of this work as follows:
\begin{itemize}
    \item We propose ICorr (in \Cref{sec:ICorr}), which enforces the correlation constraint throughout training process, and demonstrate its benefits through theoretical analysis of causality (in \Cref{sec:theory}).
    \item 
    We present the motivation of ICorr through a case study in \Cref{sec: case study}, which reveals that when in noisy environments, previous IRM-related methods may fail to get the optimal IRM solution because of environmental inherent noises, whereas ICorr can still converge to the optimal IRM solution.
    \item An extensive set of empirical results is provided to demonstrate that ICorr can generalize better in noisy environments across different datasets when compared with other domain generalization methods (\Cref{sec: experiments}).
\end{itemize}

\section{Study IRM in noisy environments}
\subsection{Preliminaries}
Given that $\mathcal{X}$ and $\mathcal{Y}$ are the input and output spaces respectively, let $\eall:=\{ e_1, e_2,...,e_m\}$ be a collection of $m$ environments in the sample space $\mathcal{X} \times \mathcal{Y}$ with different joint distributions $\pr^e(\xx^e, \yy)$, where $e \in \eall$.
Consider $\etr\subset \eall$ to be the training environments and $\St^e:=\{ (\x_i^e,\y_i) \}_{i=1}^{n^e}$ to be the training dataset drawn from distribution $\pr^e(\xx^e, \yy)$ ($e \in \etr$) with $n^e$ being dataset size. 
Given the above training datasets $\St^e$ ($e \in \etr$), the task is to learn an optimal model $f(\cdot ; \w) : \mathcal{X} \to \mathcal{Y}$, 
such that $f(\xx^e; \w)$ performs well in predicting $\yy$ when given $\xx^e$ not only for $e \in \etr$ but also for $e \in \eall \setminus\etr$, where $\w$ is the parameters of $f$.

The ERM algorithm \cite{vapnik1991principles} tries to solve the above problem via directly minimizing the loss throughout training environments:
\begin{equation}\nonumber
    \min _{\w: \mathcal{X} \rightarrow \mathcal{Y}} \sum_{e \in \mathcal{E}_{\mathrm{tr}}} \R^e(\w), \tag{ERM}
\end{equation}
where $\R^e(\w)$, $\R(\xx^e,\w)$ are the expected loss of $f(\cdot ; \w)$ in the environment $e$, the loss of $f(\xx^e ; \w)$ for the data $\xx^e$, respectively.

IRM \cite{DBLP:journals/corr/abs-1907-02893} firstly supposes that the predictor $f(\cdot ; \w)$ can be made up of $g(\cdot; \Phi)$ and $h(\cdot ; \vv)$, i.e., $f(\cdot ; \w) = h(g(\cdot ; \Phi); \vv)$,
where $\w = \{\vv, \Phi\}$ are the model parameters.
Here, $g(\cdot ; \Phi) : \mathcal{X} \to \mathcal{H}$ extracts invariant features among $\etr$ through mapping $\mathcal{X}$ to the representation space $\mathcal{H}$. 
The classifier $h(\cdot ; \vv) : \mathcal{H} \to \mathcal{Y}$ is supposed to be simultaneously optimal for all training environments. 
The original IRM method learns $g(\cdot ; \Phi)$ and $h(\cdot ; \vv)$ through solving the following minimization problem:
\begin{equation}\nonumber
\begin{aligned}
    &\mathop{\min}_{\Phi: \mathcal{X} \rightarrow \mathcal{H} \atop
    \vv: \mathcal{H} \rightarrow \mathcal{Y}}
    \sum_{e \in \mathcal{E}_{\mathrm{tr}}} \mathcal{R}^e(\{\vv,\Phi\}) \\
    & s.t. \quad \vv \in \mathop{\argmin}_{\bar\vv:\mathcal{H} \rightarrow \mathcal{Y}}\R^e(\{\bar \vv,\Phi\}), \; \mathrm{for\; all} \; e\in\etr. 
\end{aligned} \tag{IRM}
\end{equation}
However, IRM remains a bi-level optimization problem.
\citet{DBLP:journals/corr/abs-1907-02893} suggest, for practical reasons, to relax this strict limitation by using the method IRMv1 as a close approximation to IRM:
\begin{equation}\nonumber
    \min _{\w: \mathcal{X} \rightarrow \mathcal{Y}} \sum_{e \in \mathcal{E}_{\mathrm{tr}}} \Big [\R^e(\w)+\lambda \big\|\nabla_{\vv \mid \vv=1} \R^e(\w)\big\|^2\Big], \tag{IRMv1}
\end{equation}
where $\vv=1$ is a scalar and fixed ``dumm'' classifier.
Furthermore, VREx \cite{krueger2021out} adopts the following regularizer for robust optimization:
\begin{equation}\nonumber
    \min _{\w: \mathcal{X} \rightarrow \mathcal{Y}} \lambda \cdot \Var(\R^{e}(\w)) + \sum_{e \in \mathcal{E}_{\mathrm{tr}}} \R^e(\w), \tag{VREx}
\end{equation}
where $\Var(\R^{e}(\w))$ represents the variance of the losses $\R^e(\w)$ in $\etr$.
Clearly, to encourage $f(\cdot;\w)$ to be simultaneously optimum, IRMv1 constrains the gradients $\nabla_{\vv \mid \vv=1} \R^e(\w)$ to be 0 and VREx decreases the loss variance $\Var(\R^{e}(\w))$ to 0.

\subsection{Invariant correlation of representation with label}
\label{sec:ICorr}
We now formally describe our method (ICorr) to extract invariant features in noisy environments.
ICorr performs robust learning via stabilizing the correlation between representation and true label across environments:
\begin{equation}\nonumber
    \min _{\w: \mathcal{X} \rightarrow \mathcal{Y}} \lambda \cdot \mathop{\Var}(\rho^e_{f,\yy}(\w)) + \sum_{e \in \mathcal{E}_{\mathrm{tr}}} \R^e(\w), \tag{ICorr}
\end{equation}
where $\rho^e_{f,\yy}(\w)=\EE_{\xx^e,\yy}(\tilde f(\xx^e;\w)\yy)$ is the correlation between $f(\xx^e;\w)$ and $\yy$ in the environment $e$, 
$\tilde f(\xx^e;\w) = f(\xx^e;\w)-\EE_{\xx^e}(f(\xx^e;\w))$, and $\mathop{\Var}(\rho^e_{f,\yy}(\w))$ represents the variance of the correlation in $\etr$. 
Here $\lambda\in [0, +\infty)$ controls the balance between reducing average loss and enhancing stability of correlation, with $\lambda = 0$ recovering ERM, and $\lambda\to +\infty$ leading ICorr to focus entirely on making the correlation equal. 
In the following, we demonstrate the power of ICorr in noisy environments through the case study (\Cref{sec: case study}) and the theoretical analysis of causality (\Cref{sec:theory}), respectively.

\begin{figure*}[t!]
\includegraphics[width=0.95
\textwidth]{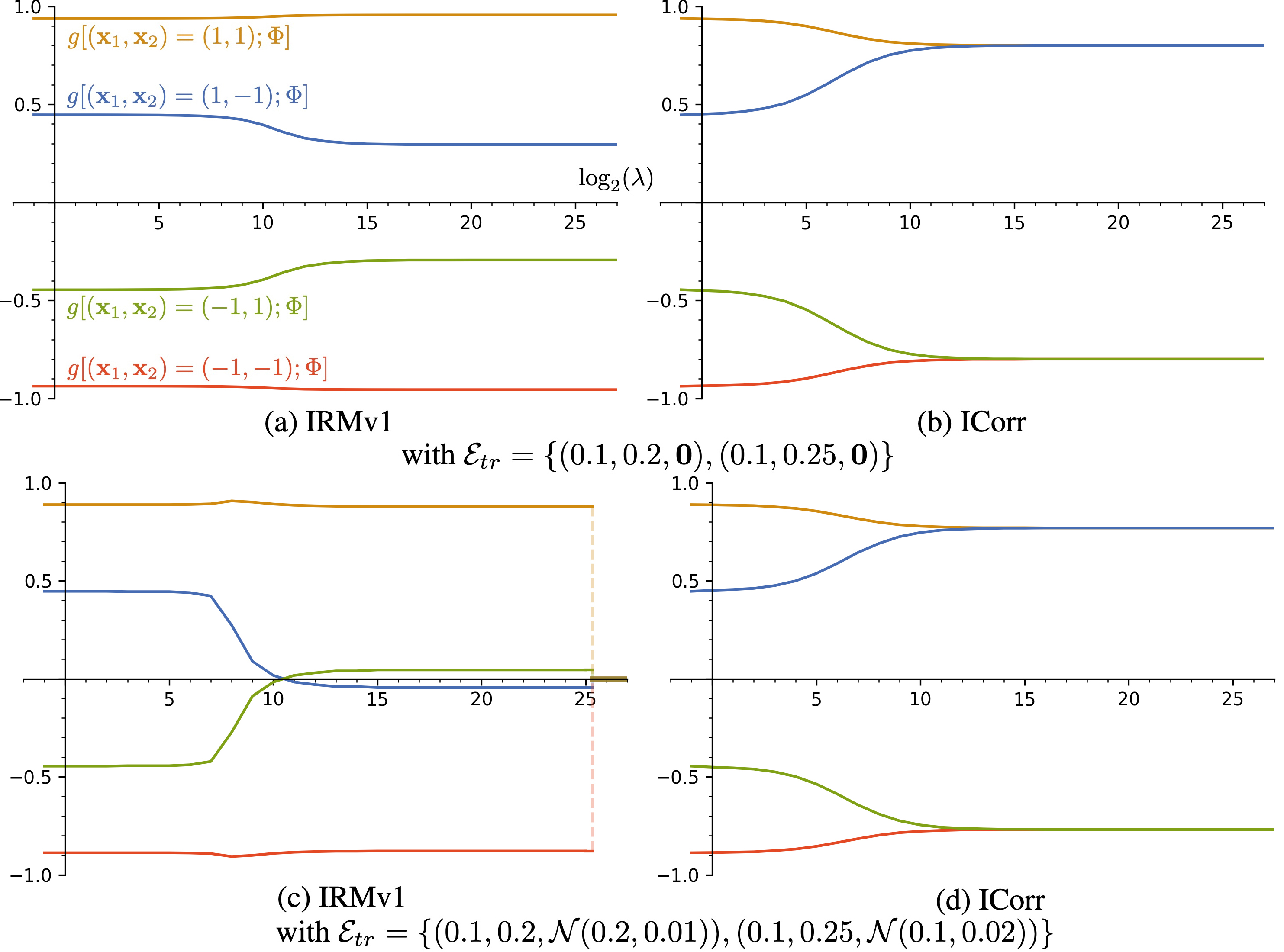}
\centering
\vspace{0mm}
\caption{
The output (vertical axis) of optimized $g(\xx^e;\Phi)$ with four inputs $(\x_1,\x_2)=\{\textcolor[RGB]{208, 138, 17}{(1,1)},\textcolor[RGB]{71, 108, 179}{(1,-1)},$ $\textcolor[RGB]{128, 161, 22}{(-1,1)},\textcolor[RGB]{227, 73, 37}{(-1,-1)}\}$. 
The horizontal axis is $\log_2(\lambda)$, with $-1$ representing $\lambda = 0$.
\textbf{(a)}, \textbf{(b)} are the results of IRMv1 and ICorr for varying $\lambda$ optimized with training environments $\etr=\{ (0.1,0.2,\0), (0.1,0.25,\0) \}$.
\textbf{(c)}, \textbf{(d)} are the results of IRMv1 and ICorr optimized with $\etr=\{ (0.1,0.2,\mathcal{N}(0.2,0.01)), (0.1,0.25,\mathcal{N}(0.1,0.02)) \}$.
More results are given in Appendix~\ref{appendix: A}.
}
\vspace{0mm}
\label{fig:1}
\end{figure*}
\subsection{Why is ICorr necessary (a case study in two-bit environments)}
\label{sec: case study}

\citet{DBLP:journals/corr/abs-1907-02893} present the Colored-MNIST task, a synthetic challenge derived from MNIST, to demonstrate the efficacy of the IRM technique and IRMv1 in particular.
Although MNIST pictures are grayscale, Colored-MNIST images are colored red or green in a manner that strongly (but spuriously) correlates with the class label.
In this case, ERM successfully learns to exploit the color during training, but it fails at test time when the correlation with the color is inverted.

\citet{pmlr-v130-kamath21a} study an abstract version of Colored-MNIST based on two bits of input, where $\yy$ is the label to be predicted, $\hat\xx_1$ is correlated with the label of the hand-written digit ($0-4$ or $5-9$), and $\hat\xx^e_2$ corresponds to the color (red or green).

\begin{table}[t]
\centering
\caption{The square losses for optimal IRM (oracle) and other optimization methods: ERM, IRMv1($\lambda=+\infty$), VREx($\lambda=+\infty$), ICorr($\lambda=+\infty$).
All losses in this table are computed with $\eta^e=\0$, \textbf{left} methods are optimized with training environments $\etr=\{ (0.1,0.2,\0), (0.1,0.25,\0) \}$, whereas \textbf{right} ones are optimized with $\etr=\{ (0.1,0.2,\mathcal{N}(0.2,0.01)), (0.1,0.25,\mathcal{N}(0.1,0.02)) \}$.
The upper two rows are the results with training $\beta^e$ ($0.2$ and $0.25$), whereas the lower two rows present the results when the correlation of $\hat\xx^e_2$ has flipped ($\beta^e = 0.7, 0.9$). 
In addition, we also provide more results of other methods in Appendix~\ref{appendix: A}.
Best results are in \textbf{bold}.}
\label{tab:1}
\vspace{0mm}
\scalebox{0.7}{
\begin{tabular}{lcccccccccccccc}
\specialrule{.1em}{.075em}{.075em} 
\multicolumn{1}{c}{\multirow{2}{*}{$\R(\alpha,\beta^e,\eta^e)$}} 
&& \multicolumn{6}{c}{$\etr=\{ (0.1,0.2,\0), (0.1,0.25,\0) \}$} && \multicolumn{6}{c}{$\etr=\{ (0.1,0.2,\mathcal{N}_{[0.2,0.01]}), (0.1,0.25,\mathcal{N}_{[0.1,0.02]}) \}$} \\
\cline{3-8} \cline{10-15}
&& Oracle && ERM & IRMv1 & VREx & ICorr & & Oracle && ERM & IRMv1 & VREx & ICorr       
\\
\cline{0-0} \cline{3-3} \cline{5-8} \cline{10-10} \cline{12-15}  
$\R(0.1,0.2,\0)$ && \textbf{0.18} && 0.15 & 0.15 & \textbf{0.18} & \textbf{0.18} && \textbf{0.1805} && 0.15 & 0.50 & 0.50 & \textbf{0.1805}    \\
$\R(0.1,0.25,\0)$ && \textbf{0.18} && 0.16 & 0.17 & \textbf{0.18} & \textbf{0.18} && \textbf{0.1805} && 0.16 & 0.50 & 0.50 & \textbf{0.1805}    \\
\cline{0-0} \cline{3-3} \cline{5-8} \cline{10-10} \cline{12-15}
$\R(0.1,0.7,\0)_{tst}$ && \textbf{0.18} && 0.26 & 0.32 & \textbf{0.18} & \textbf{0.18} && \textbf{0.1805} && 0.25 & 0.50 & 0.50 & \textbf{0.1805}    \\
$\R(0.1,0.9,\0)_{tst}$ && \textbf{0.18} && 0.30 & 0.38 & \textbf{0.18} & \textbf{0.18} && \textbf{0.1805} && 0.30 & 0.50 & 0.50 & \textbf{0.1805}    \\
\specialrule{.1em}{.075em}{.075em}
\end{tabular}
}
\vspace{0mm}
\end{table}

\noindent\textbf{Setting:} Following \citet{pmlr-v130-kamath21a}, we initially represent each environment $e$ with two parameters $\alpha, \beta^e \in[0, 1]$. 
The data generation process is then defined as
\begin{equation}
\begin{aligned}
    \text{Invariant feature: }&\hat \xx_1 \gets \rad(0.5),  \\
    \text{True label: }&\yy \gets \hat\xx_1\cdot \rad(\alpha), \\
    \text{Spurious feature: }&\hat\xx^e_2 \gets \yy \cdot \rad(\beta^e), 
\end{aligned}
\end{equation}
where $\rad(\delta)$ is a random variable taking value $-1$ with probability $\delta$ and $+1$ with probability $1-\delta$.
In addition, we also consider an environmental inherent noise $\eta^e$.
That is, we can only observe the features interfered by environmental noise:
\begin{equation}
\begin{aligned}
    \text{Observed invariant feature: }&\xx^e_1 \gets \hat\xx_1 + \eta^e,  \\
    \text{Observed spurious feature: }&\xx^e_2 \gets \hat\xx^e_2 + \eta^e, 
\end{aligned}
\end{equation}
where $\eta^e  \sim \mathcal{N}(\mu^{e},(\sigma^{e})^2)$ is an independent Gaussian noise.

Then, for convenience, we denote an environment $e$ as $(\alpha,\beta^e,\eta^e)$, where $\alpha$ represents invariant correlation between $\hat\xx_1$ and $\yy$, $\beta^e$ represents varying (non-invariant) correlation between $\hat\xx^e_2$ and $\yy$ across $\eall$,  $\eta^e$ is the environmental inherent noise.
We consider a linear model ($f(\xx^e;\w)_{\vv=1}=g(\xx^e; \Phi) = w_1\xx^e_1 + w_2\xx^e_2$) with square loss $\R_{sq}(\hat\yy, \yy) := \frac{1}{2}(\hat \yy-\yy)^2$ in this case study.
All methods are optimized in training environments $\etr=\{ (0.1,0.2,\eta^{e_1}), (0.1,0.25,\eta^{e_2}) \}$ with $\eta^{e_{1,2}}=\0$ ({Case 1}) or $\eta^{e_{1,2}}\ne \0$ ({Case 2}). 

\quad

\noindent \textbf{Case 1: Optimization without environmental inherent noise.}

In the first case, $\mathcal{E} = \mathcal{E}_{\alpha=0.1}$ with training environments $\etr = \{e_1 = (0.1, 0.2,$ $\0)$, $e_2 = (0.1, 0.25, \0)\}$,
our results are similar to \citet{pmlr-v130-kamath21a}. 

\noindent$\bullet$ \textbf{Failure of IRMv1:} Consider that IRMv1, VREx, ICorr become exactly ERM when their regularization terms are $\lambda=0$.
\Cref{fig:1}(a) shows the output of $g(\xx^e;\Phi)$ from IRMv1 ($\lambda=0$, ERM) to IRMv1 ($\lambda=+\infty$) with four inputs.
Note that IRMv1 with a specific $\lambda$ is optimized by training environments $\etr$.
We find that $g((1,-1);\Phi)$ decreases and $g((-1,1);\Phi)$ increases with growing $\lambda$; this phenomenon demonstrates the reliance on $\xx^e_2$ increases when $\lambda \to +\infty$.
Thus IRMv1 may find an un-invariant predictor even worse than ERM.
This is also echoed by the results in \Cref{tab:1}(left):
When the correlation of $\hat \xx^e_2$ has flipped ($\beta^e = 0.7, 0.9$) in the test environment, the performance of the predictor from IRMv1 ($\lambda=+\infty$) may be worse than that learnt by optimal IRM and even worse than ERM.

\noindent$\bullet$ \textbf{Success of VREx, ICorr:}
Fortunately, VREx and ICorr can still converge to the optimal IRM with increasing $\lambda$, as stabilizing losses (VREx) or correlations (ICorr) across different training environments can effectively prevent the interference from spurious feature in this case.
\Cref{fig:1}(b) demonstrates that $g(\xx^e;\Phi)$ from ICorr only relies on invariant feature $\xx^e_1$ when $\lambda \ge 2^{11}$.  
Furthermore, VREx ($\lambda=+\infty$) and ICorr ($\lambda=+\infty$) in \Cref{tab:1}(left) perform the same as optimal IRM in all training and test environments.

\begin{figure}[t]
\vspace{-4mm}
\includegraphics[width=0.95\textwidth]{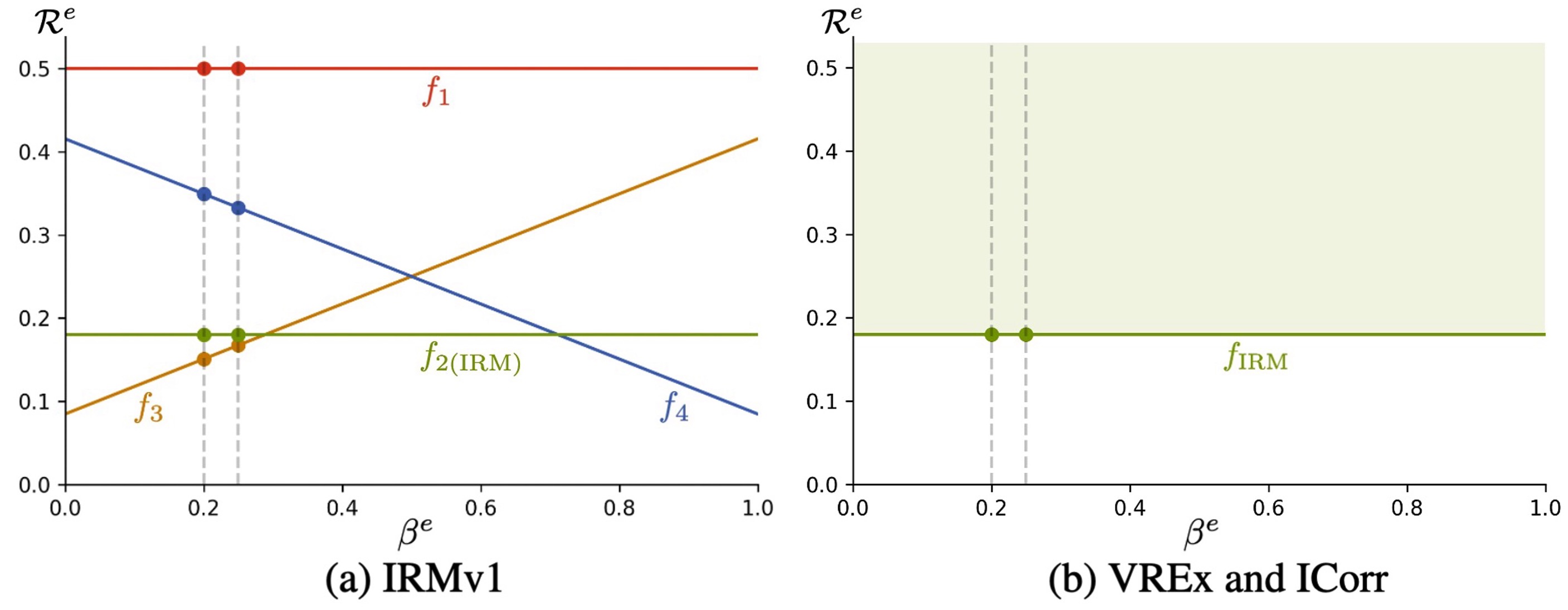}
\centering
\vspace{0mm}
\caption{The solutions for \textbf{(a)} IRMv1, \textbf{(b)} VREx and ICorr when $\lambda=+\infty$. I.e., the solutions satisfy \textbf{(a)} $\nabla_{\vv \mid \vv=1} \R^e(\w)=0$, \textbf{(b)} $\Var(\R^{e}(\w))=0$ and $\Var(\rho^e_{f,\yy}(\w))=0$ for $\etr = \{e_1 = (0.1, 0.2, \0), e_2 = (0.1, 0.25, \0)\}$.
The horizontal axis is $\beta^e$ and vertical axis represents square loss for $e=(0.1,\beta^e,\0)$. 
The solid circles are training losses for different solutions. 
Clearly, \textbf{(a)} picks $f_3$ and \textbf{(b)} picks $f_{\text{IRM}}$.
}
\vspace{-3mm}
\label{fig:2}
\end{figure}

\noindent$\bullet$ \textbf{Why:}
As shown in \Cref{fig:2}(a)~\cite{pmlr-v130-kamath21a}, there are four solutions for IRMv1 when $\lambda\to +\infty$.
Unfortunately, IRMv1 picks $f_3$ rather than optimal IRM solution ($f_2$) as $f_3$ has the lowest training loss of those four solutions.
Clearly, $f_3$ relies more on $\xx^e_2$ and damages the performance when flipping $\beta^e$. 
On the other hand, \Cref{fig:2}(b)  shows VREx and ICorr can easily converge to the optimal IRM solution when minimizing training losses for any two training environments.
The details of calculating procedure are given in Appendix~\ref{appendix:casestudy}.


\quad 

\noindent\textbf{Case 2: Optimization with environmental inherent noise.}

In the second case, we further consider training environments with environmental inherent noise, i.e., $\etr=\{ (0.1,0.2,\mathcal{N}(0.2,0.01)), (0.1,0.25,\mathcal{N}(0.1,0.02)) \}$.

\noindent$\bullet$ \textbf{Failure of IRMv1:} As shown in \Cref{fig:1}(c), compared with clean training environments in \Cref{fig:1}(a), noisy training environments may make IRMv1 more reliant on $\xx_2^e$ when $\lambda\in [2^{10},2^{25.3}]$, 
and finally IRMv1 converges to a zero solution ($w_1=0,w_2=0$) with a non-continuous step when $\lambda>2^{25.3}$.
Thus the loss for IRMv1 ($\lambda=+\infty$) in \Cref{tab:1}(right) is $0.5$ across all environments. 
This finding is consistent with our calculation in Appendix~\ref{appendix:casestudy}, which demonstrates that IRMv1 ($\lambda=+\infty$) has only one zero solution.

\noindent$\bullet$ \textbf{Failure of VREx:} In noisy training environments, VREx ($\lambda=+\infty$) in \Cref{tab:1}(right) also fails to extract invariant feature,
since minimizing $\Var(\R^{e}(\w))$ cannot help find the optimal invariant predictor when there are different environmental inherent noises. 
As shown in \Cref{fig:app1}(a) of Appendix~\ref{appendix: A}, VREx also converges to a zero solution when $\lambda \to +\infty$.

\noindent$\bullet$ \textbf{Success of ICorr:} ICorr can deal with this case as its regularization term only considers the correlation between representation and true label. 
In other words, it can filter out the impact of environmental noise which is independent of true label. 
The results in \Cref{fig:1}(d) show that ICorr still converges to IRM solution in noisy training environments and \Cref{tab:1}(right) shows that ICorr ($\lambda=+\infty$) has the same results with optimal IRM (oracle).

\begin{figure*}[t!]
\includegraphics[width=0.95
\textwidth]{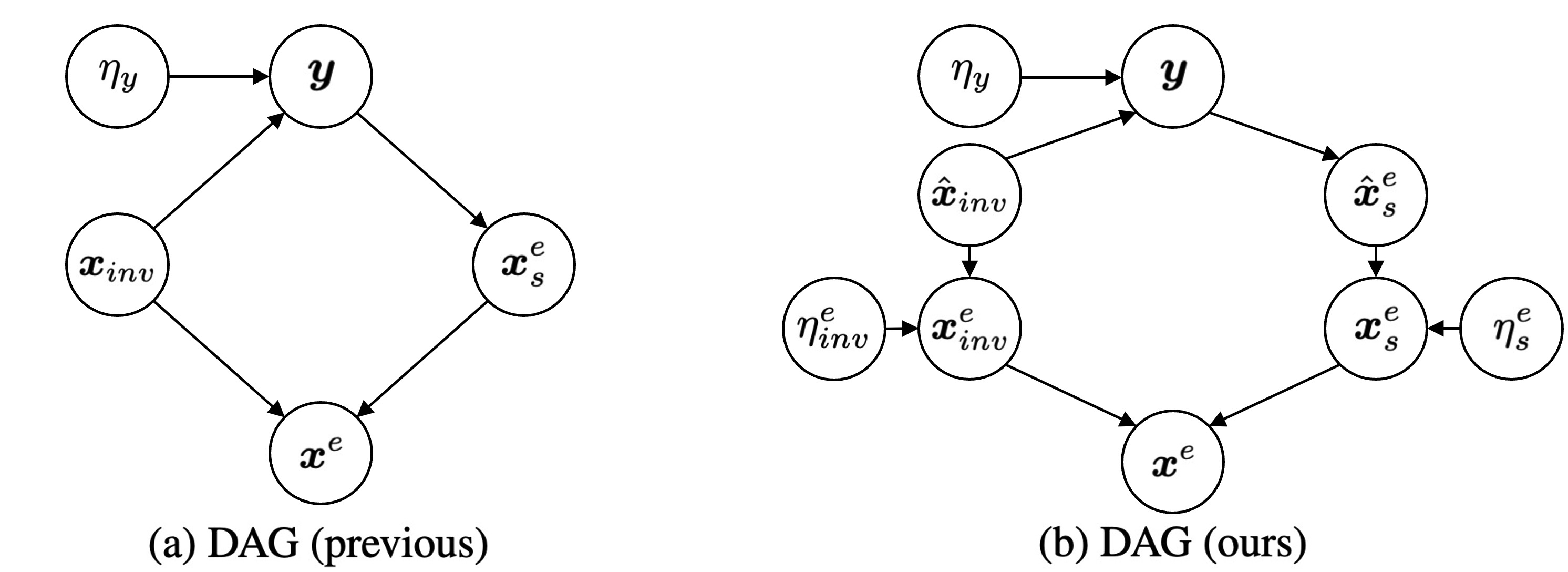}
\centering
\vspace{0mm}
\caption{Comparison of the DAG from \textbf{(a)} \cite{DBLP:journals/corr/abs-1907-02893,pmlr-v130-kamath21a} and \textbf{(b)} ours.
Different from \textbf{(a)}, the observed invariant feature $\xx_{inv}^e$ in \textbf{(b)} is affected by the environmental inherent noise $\eta_{inv}^e$, such as snow covering the face or water blurring the body in \Cref{fig: example}.
}
\vspace{-4mm}
\label{fig:3}
\end{figure*}

\noindent$\bullet$ \textbf{Why:}
Due to the variability of environmental inherent losses, 
optimizing $||\nabla_{\vv \mid \vv=1} \R^e(\w)||\to 0$ or $\Var(\R^{e}(\w))\to 0$ may be impractical in noisy training environments.
That is to say, if an optimal IRM predictor operates in noisy training environments, there may exist $||\nabla_{\vv \mid \vv=1} \R^e(\w)||\ne 0$ and $\Var(\R^{e}(\w))\ne 0$ due to different environmental inherent noises. 
Nevertheless, the independence between $\eta^e$ and $\yy$ ensures that $\mathop{\Var}(\rho^e_{f,\yy}(\w))=0$ holds for the optimal IRM predictor. 
Details of the calculation are given in Appendix~\ref{appendix:casestudy}.
(We provide formal proofs under a more general setting for the above claims in the next section.)

\noindent$\bullet$ \textbf{Failure of other methods:} 
In addition, gradient-based optimization methods for optimal IRM can also be unsuccessful in noisy environments. 
In this noisy case with $\etr=\{e_1,e_2\}$, IGA~\cite{koyama2020out} minimizes $||\nabla_{\w}\R^{e_1}(\w)-\nabla_{\w}\R^{e_2}(\w)||_2^2$, \citet{shi2021gradient} increases $\nabla_{\w}\R^{e_1}(\w)\cdot\nabla_{\w}\R^{e_2}(\w)$, AND-mask~\cite{parascandolo2020learning} and \citet{mansilla2021domain} update weights only when $\nabla_{\w}\R^{e_1}(\w)$ and $\nabla_{\w}\R^{e_2}(\w)$ point to the same direction, 
Fishr~\cite{rame2022fishr} reduces $||\Var(\nabla_{\w}\R(\xx^{e_1},\w))-\Var(\nabla_{\w}\R(\xx^{e_2},\w))||_2^2$. 
Clearly, they may be failed in noisy environments as their penalty terms are also affected by environmental inherent noises. 
We provide more simulation and calculation results for some of these methods in Appendix~\ref{appendix: A} and Appendix~\ref{appendix:casestudy others} respectively.

\section{Theoretical analysis from causal perspective}
\label{sec:theory}
In this section, we present our theoretical understanding of ICorr from the perspective of causality.
Following the theoretical setting from \citet{DBLP:journals/corr/abs-1907-02893} and \citet{peters2016causal}, we formally prove that (1) $\Var(\rho^e_{f,\yy}(\w))=0$ is a necessary condition for the optimal invariant predictor in noisy environments; 
(2) $||\nabla_{\vv \mid \vv=1} \R^e(\w)||=0$, $\Var(\R^{e}(\w))= 0$ and some other minimal penalty terms may not be necessary for the optimal invariant predictor in noisy environments.

\noindent
\textbf{Setting:} 
Consider several training environments $\etr=\{e_1,e_2,...\}$ and $\xx^e$ to be the observed input of $e\in \etr$.
We adopt an anti-causal framework \cite{DBLP:journals/corr/abs-1907-02893} with data generation process as follows:
\vspace{-1mm}
\begin{equation}\nonumber
\begin{aligned}
&\yy=\gamma^\top \hat\xx_{inv} +\eta_{y},\\
&\xx^{e}_{inv} = \hat\xx_{inv} + \eta^{e}_{inv}, \quad \xx^{e}_s = \hat\xx^{e}_s + \eta^{e}_s, \\ 
&\xx^e=S\left(\begin{smallmatrix}\xx^{e}_{inv} \\ \xx_s^{e} \end{smallmatrix}\right),
\vspace{-1mm}
\end{aligned}
\end{equation}
where $\gamma\in \RR^{d_{inv}}$ and $\gamma \ne \0$, the hidden invariant feature $\hat\xx_{inv}$ and the observed invariant feature $\xx^e_{inv}$ take values in $\RR^{d_{inv}}$, the hidden spurious feature $\hat\xx_s^{e}$ and the observed spurious feature $\xx_s^e$ take values in $\RR^{d_s}$, and $S: \RR^{(d_{inv}+d_s)}\to \RR^d$ is an inherent mapping to mix features.
The hidden spurious feature $\hat\xx_s^{e}$ is generated by $\yy$ with any \emph{non-invariant} relationship, $\eta^{e}_{inv}$ and $\eta^{e}_s$ are independent Gaussian with bounded mean and variance changed by environments, $\eta_{y}$ is an independent and invariant zero-mean Gaussian with bounded variance.   
As the directed acyclic graph (DAG) in \Cref{fig:3}(b) shows, the hidden invariant feature $\hat\xx_{inv}$ generates the true label $\yy$ and $\yy$ generates the hidden spurious feature $\hat\xx^{e}_s$.
In consideration of environmental noise, we can only observe the input $\xx^e$ which is a mixture of  $\xx^e_{inv}$ and $\xx^e_{s}$ after mapping. 
(Note that the observed feature is generated by applying environmental noise to the hidden feature.)
We aim to learn a classifier to predict $\yy$ based on $\xx^e$, i.e., $f(\xx^e;\w)=h(g(\xx^e ; \Phi); \vv)$.

Drawing upon the foundational assumption from IRM~\cite{DBLP:journals/corr/abs-1907-02893}, i.e., assume that there exists a mapping $\tilde S: \RR^d \to \RR^{d_{inv}}$ such that $\tilde S (S (\begin{smallmatrix}\x_1 \\\x_2 \end{smallmatrix})) = \x_1$ for all $\x_1 \in \RR^{d_{inv}} \text{ and } \x_2 \in \RR^{d_s}$,
the following theorem mainly states that, in noisy environments, if there exists a representation $\Phi$ that elicits the optimal invariant predictor $f(\cdot;\w)$ across all possible environments $\eall$, then the correlation between $f(\xx^e;\w)$ and $\yy$ remains invariant for all $e\in \eall$.

\begin{thm}
\label{thm1}
Assume that there exists a mapping $\tilde S: \RR^d \to \RR^{d_{inv}}$ such that $\tilde S (S (\begin{smallmatrix}\x_1 \\\x_2 \end{smallmatrix})) = \x_1$ for all $\x_1 \in \RR^{d_{inv}}, \x_2 \in \RR^{d_s}$.
Then, if $\Phi$ elicits the desired (optimal) invariant predictor $f(\cdot;\w)=\gamma^\top \tilde S (\cdot)$, we have
\begin{small}
\begin{equation}\nonumber
\rho^e_{f,\yy}(\w)=\EE_{\xx^e, \yy}[\vv^\top \Phi \xx^e \yy]-\EE[\vv^\top \Phi \xx^e] \EE[\yy]=\Var(\gamma^\top \hat\xx_{inv})
\end{equation}
\end{small}holds for all $e\in \eall$. Thus we get $\Var(\rho^e_{f,\yy}(\w))=0$.
\end{thm}
\noindent\emph{Proof.}
See Appendix~\ref{appendix:proofs}. 
\hfill $\square$

\vspace{2mm}

Theorem~\ref{thm1} indicates that in noisy environments, minimizing the regularization term of ICorr, i.e., $\Var(\rho^e_{f,\yy}(\w))$, is a necessary condition to find the invariant features.
The intuition behind Theorem~\ref{thm1} is that, the correlation between the representation and the true label can effectively prevent interference in noisy environments, whereas IRMv1 and VREx may get stuck.
In the following, we would like to point out that the regularization strategies employed in IRMV1, VREx and others may not be the most effective.

\begin{mycor}
\label{lem2}
If $\Phi$ elicits the desired invariant predictor $f(\cdot;\w)=\gamma^\top \tilde S (\cdot)$, then 
there exists $e$ satisfies
\begin{equation}\nonumber
    \frac{\partial \R^e(\w)}{\partial \vv|_{\vv=1}}\ne 0
\end{equation}
in noisy environments.
\end{mycor}
\noindent\emph{Proof.}
See Appendix~\ref{appendix:proofs}. 
\hfill $\square$
\vspace{2mm}

Corollary~\ref{lem2} suggests that $||\nabla_{\vv \mid \vv=1} \R^e(\w)||=0$ (IRMv1) may not be a necessary condition for the optimal invariant predictor in noisy environments,
as environmental inherent losses can lead to non-zero $||\nabla_{\vv \mid \vv=1} \R^e(\w)||$. 
Even in clean environments without noise, $||\nabla_{\vv \mid \vv=1} \R^e(\w)||=0$ may point to other predictors rather than the optimal invariant one (Case 1 in \Cref{sec: case study}).

\begin{mycor}
\label{lem3}
If $\Phi$ elicits the desired invariant predictor $f(\cdot;\w)=\gamma^\top \tilde S (\cdot)$, there exists $\eta_{inv}^{e_1}\ne \eta_{inv}^{e_2}$ in noisy environments $\{e_1,e_2\}$ such that  
\begin{equation}\nonumber
    \Var(\R^{e}(\w))\ne 0.
\end{equation}
\end{mycor}
\noindent\emph{Proof.}
See Appendix~\ref{appendix:proofs}. 
\hfill $\square$

\vspace{2mm}
Corollary~\ref{lem3} shows that $\Var(\R^{e}(\w))$ (REx) may also be failed to represent as an indicator for the optimal invariant predictor in noisy environments. 
Given different inherent losses across environments, it seems unreasonable to enforce all losses to be equal.
In Appendix~\ref{appendix:causality others}, we further prove that 
the regularization terms for IGA, Fishr and IB-ERM~\cite{ahuja2021invariance} may also be not necessary to find the optimal invariant predictor in such noisy situations.

In conclusion, 
in this section, we examine ICorr from a causal perspective and provide theoretical analysis that minimizing $\mathop{\Var}(\rho^e_{f,\yy}(\w))$ is a necessary condition to find the invariant features in noisy environments. 
On the other hand, IRMv1, VREx and others may be ineffective in obtaining the optimal invariant predictor due to the impact of environmental noise on their regularization terms.

\section{Experiments}
\label{sec: experiments}

In this section, we implement extensive experiments with ColoredMNIST~\cite{DBLP:journals/corr/abs-1907-02893}, Circle dataset~\cite{DBLP:conf/icml/WangHK20}, noisy DomainBed~\cite{gulrajani2020search} framework, noisy Waterbirds~\cite{wah2011caltech,zhou2017places,sagawa2019distributionally} and CelebA~\cite{liu2015faceattributes} datasets.
The first part includes comprehensive experiments on ColoredMNIST using multi-layer-perceptrons (MLP) with varying environmental noises.
In the second part, we conduct further experiments to verify the effectiveness of ICorr in more noisy environments with more architectures.

\begin{table*}[t]
\centering
\caption{Comparison of MLP on ColoredMNIST with varying training noises, i.e., first training environment without noise, second training environment with $\0$, $\mathcal{N}{(0,0.5)}$ and $\mathcal{N}{(0,1)}$, respectively. 
We repeat each experiment with 100 times and report the best, worst and average accuracies (\%) on the test environment with $\0$, $\mathcal{N}{(0,0.5)}$ and $\mathcal{N}{(0,1)}$, respectively.
Best results are in \textbf{bold}.
}
\label{tab:2}
\vspace{0mm}
\scalebox{0.75}{
\begin{tabular}{cclcccccccccccc}
\specialrule{.1em}{.075em}{.075em} 
\multicolumn{1}{c}{\multirow{2}{*}{Test noise}} &&\multicolumn{1}{c}{\multirow{2}{*}{Method}} 
&& \multicolumn{3}{c}{$\{\0$, $\0\}_{\text{train}}$} && \multicolumn{3}{c}{$\{\0$, $\mathcal{N}{(0,0.5)}\}_{\text{train}}$} && \multicolumn{3}{c}{$\{\0$, $\mathcal{N}{(0,1)}\}_{\text{train}}$}        
\\
&&&& Best & Worst & Mean && Best & Worst & Mean && Best & Worst & Mean         
\\
\cline{1-1} \cline{3-3} \cline{5-7} \cline{9-11} \cline{13-15} 
\multirow{7}*{$\0$} && ERM && 50.85 & 10.08 & 27.08 && 51.77 & 17.70 & 35.38 && 51.71 & 10.48 & 35.64     \\
&& IRMv1 && 70.12 & 63.31 & 67.46 && 50.65 & 17.36 & 36.92 && 50.42 & 10.13 & 31.19     \\ 
&& VREx && \textbf{70.84} & 64.80 & 69.02 && 58.66 & 23.50 & 43.18 && 51.69 & 14.43 & 32.98   \\
&& CLOvE && 69.07 & 41.32 & 64.97 && 34.00 & 10.61 & 15.83 && 50.61 & 10.77 & 31.41         \\  
&& Fishr && 70.48 & \textbf{66.01} & \textbf{69.07} && 50.18 & 20.50 & 36.98 && 50.87 & 9.87 & 27.01        \\
&& ICorr && 70.56 & 65.25 & 68.33 && \textbf{69.40} & \textbf{26.69} & \textbf{53.73} && \textbf{68.11} & \textbf{18.18} & \textbf{44.16}   \\
\cline{1-1} \cline{3-3} \cline{5-7} \cline{9-11} \cline{13-15} 
\multirow{7}*{$\mathcal{N}{(0,0.5)}$} && ERM && 51.44 & 22.63 & 36.11 && 51.71 & 13.18 & 32.98 && 51.63 & 12.28 & 32.94    \\
&& IRMv1 && 59.59 & 53.22 & 56.75 && 51.19 & 11.61 & 32.01 &&  50.89 & 10.28 & 29.33   \\ 
&& VREx &&  58.73 & 53.52 & 56.61 && 51.35 & 30.44 & 41.97 && 51.54 & 13.69 & 35.09     \\
&& CLOvE && 49.28 & 36.10 & 44.87 && 42.45 & 20.33 & 31.40 &&  49.76 & 23.32 & 41.37       \\  
&& Fishr && \textbf{62.64} & \textbf{57.10} & \textbf{60.54} && 51.40 & 26.28 & 38.17 &&  50.36 & 10.87 & 30.63      \\ 
&& ICorr && 59.38 & 53.02 & 56.32 && \textbf{64.66} & \textbf{35.43} & \textbf{57.17} && \textbf{67.09} & \textbf{23.96} & \textbf{49.00}    \\
\cline{1-1} \cline{3-3} \cline{5-7} \cline{9-11} \cline{13-15} 
\multirow{7}*{$\mathcal{N}{(0,1)}$} && ERM &&   50.90 & 32.70 & 42.36 && 51.54 & 20.61 & 36.94 && 50.99 & 18.31 & 35.86 \\
&& IRMv1 && 55.31 & 49.94 & 52.91 && 51.08 & 18.95 & 36.16 && 51.19 & 15.34 & 32.72  \\ 
&& VREx && 54.20 & 50.33 & 52.55 && 50.85 & 34.57 & 43.87 && 51.47 & 22.92 & 39.29   \\
&& CLOvE && 47.39 & 40.19 & 45.05 &&  46.12 & 31.23 & 39.65 && 49.83 & 33.78 & 45.29       \\  
&& Fishr && \textbf{57.76} & \textbf{53.48} & \textbf{55.81} && 51.05 & 34.63 & 42.17 && 51.15 & 17.33 & 34.90       \\ 
&& ICorr && 54.51 & 50.36 & 52.65 && \textbf{60.06} & \textbf{44.56} & \textbf{55.27} && \textbf{63.51} & \textbf{40.14} & \textbf{52.26}     \\
\specialrule{.1em}{.075em}{.075em}
\end{tabular}
}
\vspace{0mm}
\end{table*}

\subsection{MLP with ColoredMNIST}
\textbf{Training setting:}
This proof-of-concept experiment of ColoredMNIST follows the settings from~\citet{DBLP:journals/corr/abs-1907-02893,krueger2021out}. 
The MLP consists of two hidden layers with $256$ and $256$ units respectively.
Each of these hidden layers is followed by a ReLU activation function.
The final output layer has an output dimension of number of classes.
All networks are trained with the Adam optimizer, $\ell_2$ weight decay $0.001$, learning rate $0.001$, batchsize $25000$ and epoch $500$.
Note that we use the exactly same hyperparameters as \citet{DBLP:journals/corr/abs-1907-02893,krueger2021out}, only replacing the IRMv1 penalty and VREx penalty with ICorr penalty and other penalties.

\noindent\textbf{ColoredMNIST setting:}
We create three MNIST environments (two training and one test) by modifying each example as follows:
firstly, give the input a binary label $\tilde\yy$ depending on the digit: $\tilde\yy = 0$ for digits $0$ to $4$ and $\tilde\yy = 1$ for digits $5$ to $9$; secondly, define the final true label $\yy$ by randomly flipping $\tilde\yy$ with a probability $0.25$;
the third step is to randomly choose the color id $c$ by flipping $\yy$ with probability $\pr^e_{c}$, where $\pr^e_{c}$ is $0.2$ in the first environment, $0.1$ in the second environment, and $0.9$ in the test environment.
Finally, if $c$ is $1$, the image is colored in red, otherwise it is colored in green.

\noindent\textbf{Evaluating setting:} 
There are three training groups in our experiments: $\{\0,\0\}$, $\{\0,\mathcal{N}{(0,0.5)}\}$ and $\{\0,\mathcal{N}{(0,1)}\}$. 
Specifically, the first training environment is clean without noise (i.e., $\0$ across all three groups), the second training environment differs in three groups: non-noise $\0$ in the first group, noise $\mathcal{N}{(0,0.5)}$ in the second group and noise $\mathcal{N}{(0,1)}$ in the third group. 
We train each network in these three training groups respectively with $100$ times. 
Note that, following \citet{krueger2021out}, we only record the test accuracy which is less than corresponding training accuracy for each experiment.
We then report the best, average and worst performances (among $100$ runs) in the test domain with environmental noise $\0$, $\mathcal{N}{(0,0.5)}$ and $\mathcal{N}{(0,1)}$, respectively.

\noindent\textbf{Remark:} As shown in \Cref{tab:2}, there is no significant difference in the performances of IRMv1, VREx and ICorr (Fishr performs relatively better and CLOvE~\cite{wald2021calibration} performs relatively worse) when trained in clean environments (first thick column).
However, ICorr is the only method to efficiently tackle noisy training environments (second and third thick columns). 
For example, with the training group $\{\0,\mathcal{N}{(0,0.5)}\}$, ICorr can achieve $69.4\%$ best accuracy in the clean test environment, others can only get up to $58.66\%$; 
ICorr can achieve $57.17\%$ average accuracy in the $\mathcal{N}{(0,0.5)}$ noisy test environment, while VREx, Fishr and IRMv1 only get $41.97\%$, $38.17\%$ and $32.01\%$, respectively.

\subsection{More empirical results}

\begin{figure*}[t!]
\includegraphics[width=1
\textwidth]{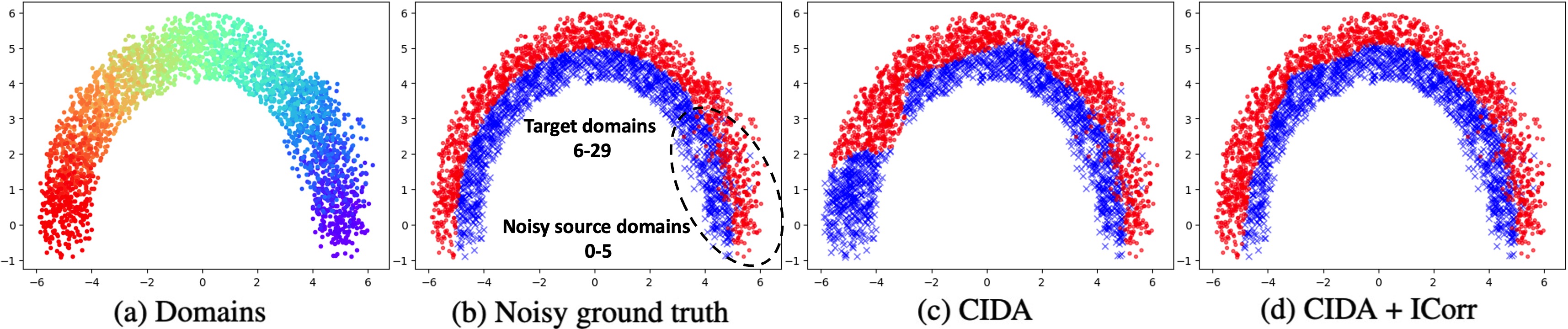}
\centering
\vspace{0mm}
\caption{
Results on the noisy circle dataset with 30 domains.
}
\vspace{-3mm}
\label{fig:5}
\end{figure*}

\begin{table}[t]
\centering
\caption{Domain generalization performances using noisy DomainBed evaluation protocol (with small environmental noises).
All methods are trained with default hyper-parameter.
We choose the checkpoint using the test domain validation set and report the corresponding test domain accuracy (\%).
}
\vspace{0mm}
\scalebox{0.95}{
\begin{tabular}{lcccccccccc}
\specialrule{.1em}{.075em}{.075em} 
&& \multicolumn{4}{c}{Noisy PACS} && \multicolumn{4}{c}{Noisy VLCS} \\
\cline{3-6} \cline{8-11}
&& A & C & P & S & & C & L & S & V  
\\
\cline{0-0} \cline{3-6} \cline{8-11}
ERM &&  86.5 & 83.9 & 94.3 & 83.4 && 97.8 & 65.1 & 69.9 & \textbf{79.2}   \\
IRMv1 && 88.1 & 85.2 & \textbf{96.4} & 73.1 && 96.1 & 67.9 & 72.1 & 77.6      \\
VREx && 86.7 & 84.3 & 95.2 & 84.8 && 96.4 & 67.4 & 73.4 & 76.4     \\
GroupDRO && 87.8 & 84.7 & 95.2 & 81.3 && 97.1 & 67.9 & 70.8 & 77.4   \\
Fishr && \textbf{89.6} & 82.0 & 94.3 & 84.9 && 98.5 & 63.2 & 70.5 & \textbf{79.2}    \\
ICorr && 89.5 & \textbf{85.5} & \textbf{96.4} & \textbf{86.4} && \textbf{98.9} & \textbf{69.6} & \textbf{73.7} & \textbf{79.2}   \\
\specialrule{.1em}{.075em}{.075em}
\end{tabular}
}
\label{tab:domainbed}
\vspace{0mm}
\end{table}

\begin{table}[t!]
\centering
\caption{The comparison of ERM, DRO~\cite{sagawa2019distributionally}, and ICorr on standard ResNet 50 with noisy Waterbirds (waterbirds vs landbirds) and noisy CelebA (blond hair vs dark hair).
We report the average accuracy (\%) across test data. 
}
\vspace{0mm}
\label{tab:a1}
\scalebox{0.95}{
\begin{tabular}{lcccc}
\specialrule{.1em}{.075em}{.075em} 
\multicolumn{1}{c}{Dataset} & & ERM & DRO & ICorr \\  
\cline{0-0} \cline{3-5}
Noisy Waterbirds (1) && 95.1 & 95.3 & \bf 96.4  \\
Noisy Waterbirds (2) && 95.5 & 95.4 & \bf 96.5   \\
\cline{0-0} \cline{3-5}
Noisy CelebA (1)  && 93.6 & 93.2 & \bf 94.3  \\
Noisy CelebA (2)  && 93.3 & 93.5 & \bf 94.1  \\
\specialrule{.1em}{.075em}{.075em} 
\end{tabular}
}
\vspace{0mm}
\end{table}

The \textbf{Circle Dataset}~\cite{DBLP:conf/icml/WangHK20} consists of $30$ domains, with indices ranging from $0$ to $29$. 
The domains are depicted in \Cref{fig:5}(a) using distinct colors (in ascending order from $0$ to $29$, from right to left). 
Each domain consists of data related to a circle, and the objective is to perform binary classification.
\Cref{fig:5}(b) illustrates the positive samples as red dots and negative samples as blue crosses.
We utilize domains $0$ to $5$ as source domains (inside dashed circle), and the remaining domains as target domains.
To create noisy environments during training, we apply Gaussian noises $\mathcal{N}(0,\text{index}/10)$ to source domains $0$ to $5$, respectively, while keeping target domains $6$ to $29$ clean.
All other settings are same with \cite{DBLP:conf/icml/WangHK20}.
As shown in \Cref{fig:5}(c), the performance of CIDA~\cite{DBLP:conf/icml/WangHK20} in noisy training environments is not good enough, but it can be improved by adding the ICorr penalty term as depicted in \Cref{fig:5}(d).

To further substantiate the effectiveness of ICorr, we conduct an evaluation within the \textbf{DomainBed}~\cite{gulrajani2020search} framework with two datasets: noisy PACS~\cite{li2017deeper} and noisy VLCS~\cite{fang2013unbiased}. 
In these datasets, we introduce environmental noise in the form of small Gaussian perturbations, denoted as $\mathcal{N}(0, i/5)$, where $i$ represents the index of the respective environment.
As shown in \Cref{tab:domainbed}, 
our findings reveal that ICorr consistently exhibits marked improvements in noisy environments when contrasted with other methods such as ERM, IRMv1, VREx, GroupDRO~\cite{sagawa2019distributionally}, and Fishr.
Although Fishr demonstrates superior performance in the first environment of noisy PACS, the discernible accuracy difference between ICorr and Fishr is minimal, amounting to merely $0.1\%$.
In all other environments across the two datasets, ICorr consistently delivers the highest level of performance.

We provide experimental results on ResNet 50 with noisy {\bf Waterbirds} and noisy {\bf CelebA} datasets in \Cref{tab:a1}, which further demonstrate the effectiveness of ICorr in deep neural networks.
In \Cref{tab:a1}, Noisy Waterbirds (1) represents applying the Gaussian noise $\mathcal{N}(0,0.2)$ to the training waterbird, $\mathcal{N}(0,0.4)$ to the training landbird, and $\mathcal{N}(0,0.6)$ to the test waterbird and test landbird, noisy Waterbirds (2) reverses this noise application scheme.
Noisy CelebA (1) represents applying $\mathcal{N}(0,0.2)$ to the training blond hair, $\mathcal{N}(0,0.4)$ to the training dark hair, and $\mathcal{N}(0,0.6)$ to the test blond and test dark hair, noisy CelebA (2) reverses this noise application scheme.
More empirical results and details are given in Appendix~\ref{appendix:experiments}.

\section{Related work}

The domain generalization problem is initially explicitly described by \citet{NIPS2011_b571ecea} and then defined by \citet{muandet2013domain}, which takes into account the potential of the target data being unavailable during model training. 
A large body of literature seeks to address the domain generalization challenge, typically through additional regularizations of ERM \cite{vapnik1991principles}.
The regularizations from \citet{motiian2017unified,namkoong2016stochastic,sagawa2019distributionally} enhance model robustness against minor distributional perturbations in the training distributions, some works \cite{zhang2022correct,liu2021just,yao2022improving} further improve this robustness with extra assumptions, while some other regularizations \cite{ganin2016domain,sun2016deep,li2018deep,li2018domain,dou2019domain,zhao2019learning} promote domain invariance of learned features.

Domain generalization can also be improved  by model averaging~\cite{cha2021swad,DBLP:conf/nips/ArpitWZX22,jin2020does}, training a model guided by meta learning~\cite{DBLP:conf/nips/RobeyPH21,li2018learning,li2019episodic,li2019feature,balaji2018metareg}, sample selection~\cite{kahng2023domain}, balanced mini-batch sampling~\cite{wang2023causal}, and indirection representations~\cite{pham2023improving}.
Additionally, by training the model on a variety of produced novel domains, data augmentation-based approaches can also increase the generalization ability, e.g. using domain synthesis to create new domains~\cite{DBLP:conf/eccv/ZhouYHX20}.
Some works also utilize the robust gradient direction to perturb data and obtained a new dataset to train the model ~\cite{shankar2018generalizing,wang2020learning,wang2023pgrad}. 
\citet{carlucci2019domain,volpi2018generalizing} construct a new dataset by solving the jigsaw puzzle. 
\citet{lee2023diversify} improve domain generalization through finding a diverse set of hypotheses and choosing the best one.
\citet{kaur2023modeling} develop the technique of causally adaptive constraint minimization to improve domain generalization.
\citet{huang2023harnessing} propose HOOD method that can leverage the content and style from each image instance to identify benign and malign (out of distribution) data.
\citet{xu2021fourier} develop a novel Fourier-based data augmentation strategy, which linearly interpolates between the amplitude spectrums of two images, to improve domain generalization.


In addition, there has been a growing trend towards integrating the principle of causal invariance \cite{pearl2009causality,louizos2017causal,goudet2018learning,ke2019learning,scholkopf2021toward} into representation learning \cite{peters2016causal,DBLP:journals/corr/abs-1907-02893,creager2021environment,parascandolo2020learning,wald2021calibration,ahuja2021invariance}.
In this context, the IRM~\cite{DBLP:journals/corr/abs-1907-02893} approach has been proposed to extract features that remain consistent across various environments, following the invariance principle introduced in \citet{peters2016causal}.
As of late, there have been several IRM-related methods developed in the community.
\citet{ahuja2020invariant} offer novel perspectives through the incorporation of game theory and regret minimization into invariant risk minimization.
\citet{ahuja2021invariance} propose to combine the information bottleneck constraint with invariance to address the case in which the invariant features capture all the information of the label.
\citet{zhou2022sparse} study IRM for overparameterized models.
\citet{ahuja2020empirical,liu2021kernelized}  endeavor to learn invariant features when explicit environment indices are not provided.
\citet{chen2022iterative} suggest utilizing the inherent low-dimensional structure of spurious features to recognize invariant features in logarithmic environments.
\citet{rosenfeld2020risks} study IRM in the non-linear regime and finds it can fail catastrophically.
\citet{pmlr-v130-kamath21a} analyze the success and failure cases of IRMv1 in clean environments.
\citet{zhang2022rich} propose constructing diverse initializations to stabilize domain generalization performance under the trade-off between ease of optimization and robust of domain generalization. 
\citet{yu2022regularization} propose a Lipschitz regularized IRM-related method to alleviate the influence of low quality data at both the sample level and the domain level. 
\citet{lu2021nonlinear} study IRM and obtain generalization guarantees in the nonlinear setting.
\citet{choe2020empirical} take an empirical study of IRMv1 across various environments by examining the performance of IRMv1 in different frameworks including text classification models and then \citet{sonar2021invariant} extend the IRM to the reinforcement learning task. \citet{mitrovic2020representation} propose a self-supervised setup method to learn the optimal representation by augmenting the data to build the second domain.
\citet{sun2023rethinking} study the generalization issue of face anti-spoofing models through IRM.
\citet{shao2022log} show that active model adaptation could achieve both good performance and robustness based on the IRM principle.
\citet{wad2022equivariance} propose a class-wise IRM method that tackles the challenge of missing environmental annotation.
\citet{lin2022bayesian} introduce Bayesian inference into IRM to its performance on DNNs.

In contrast to prior research, this paper investigates IRM in noisy environments where environmental noises can corrupt invariant features.
As a result, previous IRM-related approaches may not be effective in such scenarios.
Nevertheless, our ICorr technique can successfully handle noisy cases by utilizing the principle that the correlation of invariant representation with label is invariant across noisy environments.

\vspace{-2mm}
\section{Limitation}
\vspace{-2mm}
Our theoretical results are based on the assumption that there exists $\tilde S\in\RR^{d_{inv}\times d} $ such that $\tilde SS(\begin{smallmatrix}\x_1 \\\x_2 \end{smallmatrix}) = \x_1$, for all $\x_1 \in \RR^{d_{inv}}, \x_2 \in \RR^{d_s}$, which has also been utilized in the pioneering work of IRM by \citet{DBLP:journals/corr/abs-1907-02893}.
Nonetheless, this $\tilde S$ may not exist in DNNs when facing complicated $S$.
Although this may pose new challenges, we aim to advance our research by studying more possible cases of $S$ and $\tilde S$ in the future.

\vspace{-2mm}
\section{Conclusion}
\vspace{-2mm}

In this work, we introduce an IRM-related method named ICorr, which leverages the correlation between representation and label to overcome the challenge of training an invariant predictor in noisy environments. 
A detailed case study involving two-bit environments is conducted to elucidate why conventional methods might falter, whereas ICorr maintains its efficacy in such noisy settings. 
Through rigorous theoretical analyses of causality, we demonstrate the critical importance of maintaining invariant correlation across noisy environments to achieve the optimal IRM solution.
In addition to our theoretical insights, we conduct extensive experiments which show the superior performance of ICorr compared to other methods in such noisy cases.

\noindent \textbf{Data availability statement:} The research conducted in this work solely utilizes publicly available datasets, the code is available in the uploaded file.

\begin{appendices}

\section{More case study results}
\label{appendix: A}

\begin{figure*}[h!]
\includegraphics[width=1
\textwidth]{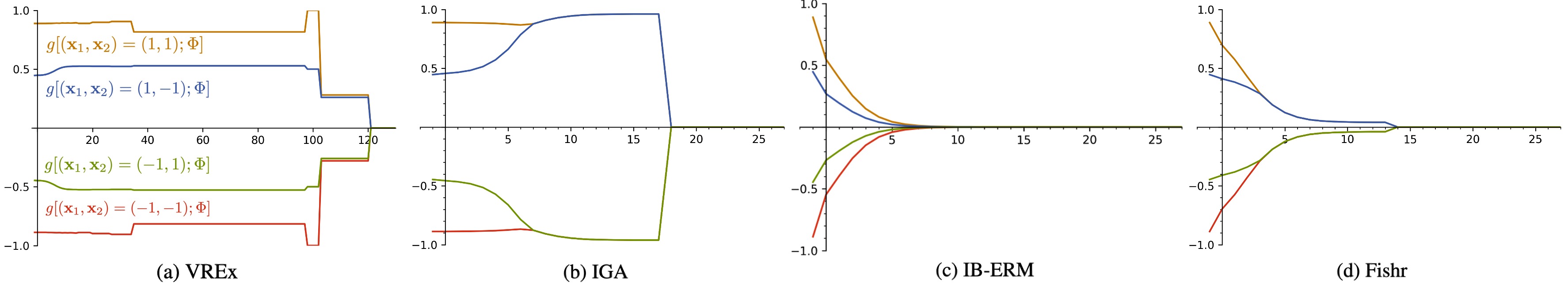}
\centering
\vspace{0mm}
\caption{
The output (vertical axis) of optimized $g(\xx^e;\Phi)$ with four inputs $(\x_1,\x_2)=\{\textcolor[RGB]{208, 138, 17}{(1,1)},\textcolor[RGB]{71, 108, 179}{(1,-1)},$ $\textcolor[RGB]{128, 161, 22}{(-1,1)},\textcolor[RGB]{227, 73, 37}{(-1,-1)}\}$. 
The horizontal axis is $\log_2(\lambda)$, with $-1$ representing $\lambda = 0$.
\textbf{(a)}, \textbf{(b)}, \textbf{(c)}, \textbf{(d)} are the results of VREx, IGA, IB-ERM and Fishr for varying $\lambda$ optimized with training environments $\etr=\{ (0.1,0.2,\mathcal{N}(0.2,0.01)), (0.1,0.25,\mathcal{N}(0.1,0.02)) \}$.
Note that in \textbf{(a)} we let $\lambda=+\infty$ when $\lambda>2^{120}$ due to numerical problems.
}
\vspace{0mm}
\label{fig:app1}
\end{figure*}

\begin{figure*}[t]
\includegraphics[width=1
\textwidth]{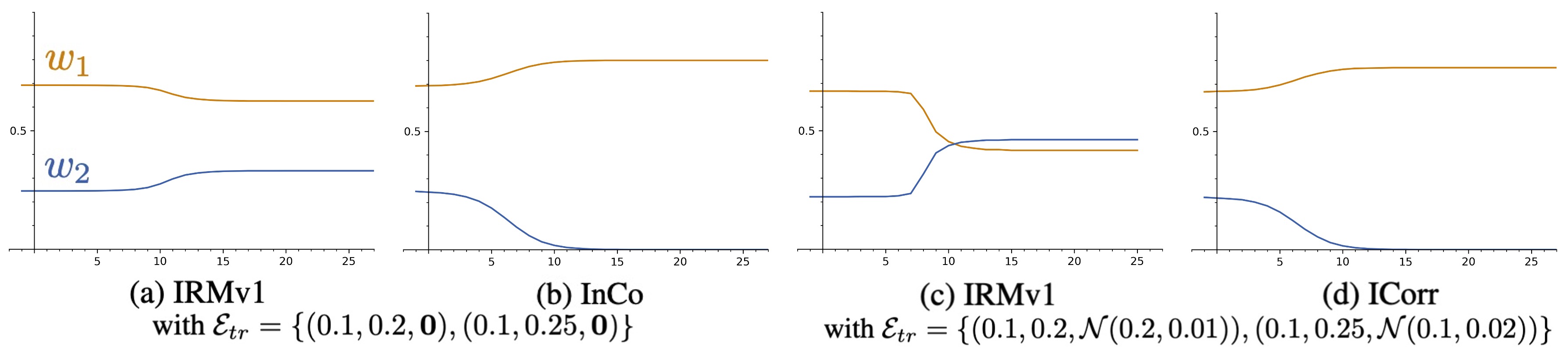}
\centering
\vspace{0mm}
\caption{
The vertical axis is the value of $\textcolor[RGB]{208, 138, 17}{w_1}$ and $\textcolor[RGB]{71, 108, 179}{w_2}$ for optimized $g(\xx^e;\Phi)$. 
The horizontal axis is $\log_2(\lambda)$, with $-1$ representing $\lambda = 0$.
\textbf{(a)}, \textbf{(b)} are the results of IRMv1 and ICorr for varying $\lambda$ optimized with training environments $\etr=\{ (0.1,0.2,\0), (0.1,0.25,\0) \}$.
\textbf{(c)}, \textbf{(d)} are the results of IRMv1 and ICorr optimized with $\etr=\{ (0.1,0.2,\mathcal{N}(0.2,0.01)), (0.1,0.25,\mathcal{N}(0.1,0.02)) \}$.
}
\label{fig:app2}
\vspace{-3mm}
\end{figure*}

\begin{table*}[h]
\centering
\caption{The square losses for optimal IRM (oracle) and different optimization methods: IGA($\lambda=+\infty$ and $2^7$), Fishr($\lambda=+\infty$ and $2^4$), IB-ERM($\lambda=+\infty$).
All losses in this table are computed with $\eta^e=\0$, and all methods are optimized with $\etr=\{ (0.1,0.2,\mathcal{N}(0.2,0.01)), (0.1,0.25,\mathcal{N}(0.1,0.02)) \}$.
The upper two rows are the results with training $\beta^e$ ($0.2$ and $0.25$), whereas the lower two rows present the results when the correlation of $\hat\xx^e_2$ has flipped ($\beta^e = 0.7, 0.9$).
}
\label{tab:app1}
\vspace{0mm}
\scalebox{0.9}{
\begin{tabular}{lcccccccc}
\specialrule{.1em}{.075em}{.075em} 
\multicolumn{1}{c}{\multirow{2}{*}{$\R(\alpha,\beta^e,\eta^e)$}} 
&& \multicolumn{7}{c}{$\etr=\{ (0.1,0.2,\mathcal{N}_{[0.2,0.01]}), (0.1,0.25,\mathcal{N}_{[0.1,0.02]}) \}$} \\
\cline{3-9}
&& Oracle && IGA & IGA($\lambda=2^7$) & Fishr & Fishr($\lambda=2^4$) & IB-ERM       
\\
\cline{0-0} \cline{3-3} \cline{5-9}  
$\R(0.1,0.2,\0)$ && \textbf{0.1805} && 0.50 & 0.36 & 0.50 & 0.40 & 0.50      \\
$\R(0.1,0.25,\0)$ && \textbf{0.1805} && 0.50 & 0.36 & 0.50 & 0.40 & 0.50      \\
\cline{0-0} \cline{3-3} \cline{5-9}
$\R(0.1,0.7,\0)_{tst}$ && \textbf{0.1805} && 0.50 & 0.36 & 0.50 & 0.40 & 0.50     \\
$\R(0.1,0.9,\0)_{tst}$ && \textbf{0.1805} && 0.50 & 0.36 & 0.50 & 0.40 & 0.50      \\
\specialrule{.1em}{.075em}{.075em}
\end{tabular}
}
\end{table*}

\begin{table*}[t]
\centering
\caption{The square losses for optimal IRM (oracle) and other optimization methods: ERM, IRMv1($\lambda=+\infty$), VREx($\lambda=+\infty$), ICorr($\lambda=+\infty$).
All losses in this table are computed with (\textbf{left}) $\eta^e=\mathcal{N}(0.2,0.01)$ and (\textbf{right}) $\eta^e=\mathcal{N}(0.1,0.02)$, all methods are optimized with $\etr=\{ (0.1,0.2,\mathcal{N}(0.2,0.01)), (0.1,0.25,\mathcal{N}(0.1,0.02)) \}$.
The upper two rows are the results with training $\beta^e$ ($0.2$ and $0.25$), whereas the lower two rows present the results when the correlation of $\hat\xx^e_2$ has flipped ($\beta^e = 0.7, 0.9$).}
\label{tab:app2}
\vspace{0mm}
\scalebox{0.68}{
\begin{tabular}{lcccccccccccccc}
\specialrule{.1em}{.075em}{.075em} 
\multicolumn{1}{c}{\multirow{2}{*}{$\R(\alpha,\beta^e,\eta^e)$}} 
&& \multicolumn{6}{c}{$\eta^e=\mathcal{N}(0.2,0.01)$} && \multicolumn{6}{c}{$\eta^e=\mathcal{N}(0.1,0.02)$} \\
\cline{3-8} \cline{10-15}
&& Oracle && ERM & IRMv1 & VREx & ICorr & & Oracle && ERM & IRMv1 & VREx & ICorr  
\\
\cline{0-0} \cline{3-3} \cline{5-8} \cline{10-10} \cline{12-15}  
$\R(0.1,0.2,\eta^e)$ && \textbf{0.1953} && 0.17 & 0.50 & 0.50 & \textbf{0.1953} && \textbf{0.1894} && 0.16 & 0.50 & 0.50 & \textbf{0.1894}    \\
$\R(0.1,0.25,\eta^e)$ && \textbf{0.1953} && 0.18 & 0.50 & 0.50 & \textbf{0.1953} && \textbf{0.1894} && 0.17 & 0.50 & 0.50 & \textbf{0.1894}    \\
\cline{0-0} \cline{3-3} \cline{5-8} \cline{10-10} \cline{12-15}
$\R(0.1,0.7,\eta^e)_{tst}$ && \textbf{0.1953} && 0.27 & 0.50 & 0.50 & \textbf{0.1953} && \textbf{0.1894} && 0.27 & 0.50 & 0.50 & \textbf{0.1894}    \\
$\R(0.1,0.9,\eta^e)_{tst}$ && \textbf{0.1953} && 0.32 & 0.50 & 0.50 & \textbf{0.1953} && \textbf{0.1894} && 0.31 & 0.50 & 0.50 & \textbf{0.1894}    \\
\specialrule{.1em}{.075em}{.075em}
\end{tabular}
}
\vspace{-3mm}
\end{table*}

As shown in \Cref{fig:app1}, we present the output of $g(\xx^e;\Phi)$ which is optimized in noisy training environments $\etr=\{ (0.1,0.2,\mathcal{N}(0.2,0.01)), (0.1,0.25,\mathcal{N}(0.1,0.02)$ $) \}$ with varying $\lambda$.
\textbf{(a)} and \textbf{(c)} show that VREx and IB-ERM converge to zero solutions when $\lambda \to +\infty$.
The results of IGA and Fishr are presented in \textbf{(b)} and \textbf{(d)}, respectively. 
Both methods converge to invariant solutions when $\lambda\ge 2^7$ for IGA and $\lambda\ge 2^4$ for Fishr, and finally they also achieve zero solutions.
However, as shown in \Cref{tab:app1}, these invariant solutions for IGA ($\lambda=2^7$) and Fishr ($\lambda=2^4$) are not optimal, as optimal loss is $0.1805$ but IGA($\lambda=2^7$) and Fishr($\lambda=2^4$) only get $0.36$ and $0.40$ respectively.
Note that here we choose $2^7$ for IGA and $2^4$ for Fishr because they are the best $\lambda$ for corresponding invariant solutions.
Fortunately, the results in \Cref{tab:1} and \Cref{fig:1}(d) demonstrate the effectiveness of ICorr to achieve optimal IRM solution (oracle) in this noisy case,
because ICorr can protect the training procedure from environmental noises.
Note that all of these simulation results are consistent with our calculation in Appendix~\ref{appendix:casestudyall}.
In \Cref{fig:app2}, we show the change of $w_1$ and $w_2$ with respect to $\lambda$.

\section{Calculation details}
\label{appendix:casestudyall}

\subsection{Calculation for IRMv1, VREx and ICorr}
\label{appendix:casestudy}

Following \citet{pmlr-v130-kamath21a} and Léon Bottou,
we provide the calculation details of IRMv1, VREx and ICorr solutions as follows.

Suppose $\etr$ consists of two environments $e_1 = (\alpha, \beta^{e_1}, \eta^{e_1})$ and $e_2 = (\alpha, \beta^{e_2},$ $\eta^{e_2})$. 
From the definition of IRMv1, VREx, ICorr, for any $f(\xx^e) = 1 \cdot g(\xx^e;\Phi) = w_1\xx^e_1 + w_2\xx^e_2$ with square loss, we have that:

when optimizing IRMv1 till $\nabla_{\vv \mid \vv=1} \R^e(\w)=0$, we get
\begin{equation}\label{eq:a4}
\begin{aligned}
&\mathbb{E}_{\xx^{e_1},\yy}\left(w_1 \xx^{e_1}_1+w_2 \xx^{e_1}_2-\yy\right)\left(w_1 \xx^{e_1}_1+w_2 \xx^{e_1}_2\right)=0, \\
&\mathbb{E}_{\xx^{e_2},\yy}\left(w_1 \xx^{e_2}_1+w_2 \xx^{e_2}_2-\yy\right)\left(w_1 \xx^{e_2}_1+w_2 \xx^{e_2}_2\right)=0;
\end{aligned}
\end{equation}

when optimizing VREx till $\Var(\R^{e}(\w))=0$, we get
\begin{equation}\label{eq:a5}
\begin{aligned}
&\mathbb{E}_{\xx^{e_1},\yy}\left(w_1 \xx^{e_1}_1+w_2 \xx^{e_1}_2-\yy\right)^2=\mathbb{E}_{\xx^{e_2},\yy}\left(w_1 \xx^{e_2}_1+w_2 \xx^{e_2}_2-\yy\right)^2;
\end{aligned}
\end{equation}

when optimizing ICorr with $\mathop{\Var}[\rho^e_{f,\yy}(\w)]=0$, we get
\begin{equation}\label{eq:a6}
\begin{aligned}
\mathbb{E}_{\xx^{e_1},\yy}(w_1 \xx^{e_1}_1\yy+w_2 \xx^{e_1}_2\yy)
=\mathbb{E}_{\xx^{e_2},\yy}(w_1 \xx^{e_2}_1\yy+w_2 \xx^{e_2}_2\yy).
\end{aligned}
\end{equation}

\noindent\textbf{Case 1}:  For both $\eta^{e_1}=\0$ and $\eta^{e_2}=\0$, 
we have ($\romannumeral1$) $\EE[(\xx^e_1)^2]=\EE[(\xx^{e}_2)^2]=1$,
($\romannumeral2$) $\EE(\xx^{e_i}_1\yy)=a$, $\EE(\xx^{e_i}_2\yy)=b_i$,
($\romannumeral3$) $\EE(\xx^{e_i}_1 \xx^{e_i}_2)=ab_i$,
where $a:=1-2\alpha$ and $b_i :=1-2\beta^{e_i}$ for $i\in \{1,2\}$.

Then, according to \eqref{eq:a4}, the solutions for IRMv1 ($\lambda=+\infty$) are
\begin{equation}\nonumber
\begin{aligned}
&(1) \quad w_1=0, w_2=0; \\
&(2) \quad w_1=a, w_2=0; \\
&(3) \quad w_1 = \frac{1}{2a}, w_2 =\sqrt{\frac{1}{2}-\frac{1}{4a^2}}, \quad s.t. \quad a^2>\frac{1}{2}; \\
&(4) \quad w_1 = \frac{1}{2a}, w_2 =-\sqrt{\frac{1}{2}-\frac{1}{4a^2}}, \quad s.t. \quad a^2>\frac{1}{2}, \; w_2\ne 0.
\end{aligned}
\end{equation}

According to (\ref{eq:a5}), the solutions for VREx ($\lambda=+\infty$) are:
\begin{equation}\nonumber
\begin{aligned}
&(1) \quad w_1=\frac{1}{a}, w_2\in \RR; \\
&(2) \quad w_1 \in \RR, w_2=0.
\end{aligned}
\end{equation}

According to (\ref{eq:a6}), the solution for ICorr ($\lambda=+\infty$) is
\begin{equation}\nonumber
\begin{aligned}
& \quad w_1 \in \RR, w_2=0.
\end{aligned}
\end{equation}

\noindent\textbf{Case 2}: $\eta^{e_1}$ and $\eta^{e_2}$ are independent but not identically distributed, i.e., $\eta^{e_1} \sim \mathcal{N}(\mu_1, \sigma_1^2)$ and $\eta^{e_2} \sim \mathcal{N}(\mu_2, \sigma_2^2)$, 
we have ($\romannumeral1$) $\EE[(\xx^{e_i}_1)^2]=\EE[(\xx^{e_i}_2)^2]=1+\mu_i^2+\sigma_i^2$,
($\romannumeral2$) $\EE(\xx^{e_i}_1\yy)=a$, $\EE(\xx^{e_i}_2\yy)=b_i$,
($\romannumeral3$) $\EE(\xx^{e_i}_1 \xx^{e_i}_2)=ab_i+\mu_i^2+\sigma_i^2$,
where $a:=1-2\alpha$ and $b_i =1-2\beta^{e_i}$ for $i\in \{1,2\}$.

According to (\ref{eq:a4}), we can calculate the solution for IRMv1 ($\lambda=+\infty$) is 
\begin{equation}\nonumber
\begin{aligned}
& w_1=0, w_2=0. 
\end{aligned}
\end{equation}

According to (\ref{eq:a5}), we can calculate the solution for VREx ($\lambda=+\infty$) is
\begin{equation}\nonumber
\begin{aligned}
& w_1=0, w_2=0. 
\end{aligned}
\end{equation}

According to (\ref{eq:a6}), the solution for ICorr ($\lambda=+\infty$) is
\begin{equation}\nonumber
\begin{aligned}
& w_1 \in \RR, w_2=0.
\end{aligned}
\end{equation}

These calculation results are also consistent with the simulations in 
\Cref{sec: case study} and Appendix~\ref{appendix: A}.

\subsection{More calculation results}
\label{appendix:casestudy others}

When optimizing IGA with $||\nabla_{\w}\R^{e_1}(\w)-\nabla_{\w}\R^{e_2}(\w)||_2^2\to 0$, we get
\begin{equation}
\begin{aligned}
&\big(\mathbb{E}_{\xx^{e_1},\yy}((\xx^{e_1}_1)^2 w_1 + w_2 \xx^{e_1}_1\xx^{e_1}_2 - \xx^{e_1}_1\yy)
-\mathbb{E}_{\xx^{e_2},\yy}((\xx^{e_2}_1)^2 w_1 +  w_2 \xx^{e_2}_1\xx^{e_2}_2 - \xx^{e_2}_1\yy)\big)^2 + ,\\
&\big(\mathbb{E}_{\xx^{e_1},\yy}((\xx^{e_1}_2)^2 w_2 + w_1 \xx^{e_1}_1\xx^{e_1}_2 - \xx^{e_1}_2\yy)
- \mathbb{E}_{\xx^{e_2},\yy}((\xx^{e_2}_2)^2 w_2 + w_1 \xx^{e_2}_1\xx^{e_2}_2 - \xx^{e_2}_2\yy)\big)^2 \to 0;
\end{aligned}
\end{equation}
when optimizing Fishr with $||\Var(\nabla_{\w}\R(\xx^{e_1},\w))-\Var(\nabla_{\w}\R(\xx^{e_2},\w))||_2^2\to 0$, we have
\begin{equation}
\begin{aligned}
&\mathbb{E}_{\xx^{e_1},\yy}((\xx^{e_1}_1)^2 w_1 + w_2 \xx^{e_1}_1\xx^{e_1}_2 - \xx^{e_1}_1\yy-\mathbb{E}_{\xx^{e_1},\yy}((\xx^{e_1}_1)^2 w_1 + w_2 \xx^{e_1}_1\xx^{e_1}_2 - \xx^{e_1}_1\yy))^2\\
&-\mathbb{E}_{\xx^{e_2},\yy}((\xx^{e_2}_1)^2 w_1 +  w_2 \xx^{e_2}_1\xx^{e_2}_2 - \xx^{e_2}_1\yy - \mathbb{E}_{\xx^{e_2},\yy}((\xx^{e_2}_1)^2 w_1 +  w_2 \xx^{e_2}_1\xx^{e_2}_2 - \xx^{e_2}_1\yy))^2 \to 0,\\
&\mathbb{E}_{\xx^{e_1},\yy}((\xx^{e_1}_2)^2 w_2 + w_1 \xx^{e_1}_1\xx^{e_1}_2 - \xx^{e_1}_2\yy-\mathbb{E}_{\xx^{e_1},\yy}((\xx^{e_1}_2)^2 w_2 + w_1 \xx^{e_1}_1\xx^{e_1}_2 - \xx^{e_1}_2\yy))^2\\
&-\mathbb{E}_{\xx^{e_2},\yy}((\xx^{e_2}_2)^2 w_2 +  w_1 \xx^{e_2}_1\xx^{e_2}_2 - \xx^{e_2}_2\yy - \mathbb{E}_{\xx^{e_2},\yy}((\xx^{e_2}_2)^2 w_2 +  w_1 \xx^{e_2}_1\xx^{e_2}_2 - \xx^{e_2}_2\yy))^2 \to 0;
\end{aligned}
\end{equation}
when optimizing IB-ERM till $\Var(g(\xx^{e_1};\Phi)|y=1)+\Var(g(\xx^{e_2};\Phi)|y=1)+\Var(g(\xx^{e_1};\Phi)|y=-1)+\Var(g(\xx^{e_2};\Phi)|y=-1)=0$, we can get
\begin{equation}
\begin{aligned}
&\mathbb{E}_{\xx^{e_1}}(w_1\xx^{e_1}_1 + w_2\xx^{e_1}_2 -\mathbb{E}_{\xx^{e_1}}(w_1\xx^{e_1}_1 + w_2\xx^{e_1}_2)|y=1)^2=0, \\
&\mathbb{E}_{\xx^{e_2}}(w_1\xx^{e_2}_1 + w_2\xx^{e_2}_2 -\mathbb{E}_{\xx^{e_2}}(w_1\xx^{e_2}_1 + w_2\xx^{e_2}_2)|y=1)^2=0, \\
&\mathbb{E}_{\xx^{e_1}}(w_1\xx^{e_1}_1 + w_2\xx^{e_1}_2 -\mathbb{E}_{\xx^{e_1}}(w_1\xx^{e_1}_1 + w_2\xx^{e_1}_2)|y=-1)^2=0, \\
&\mathbb{E}_{\xx^{e_2}}(w_1\xx^{e_2}_1 + w_2\xx^{e_2}_2 -\mathbb{E}_{\xx^{e_2}}(w_1\xx^{e_2}_1 + w_2\xx^{e_2}_2)|y=-1)^2=0.
\end{aligned}
\end{equation}

Given \textbf{case 2}, the solutions for IGA ($\lambda=+\infty$), Fishr ($\lambda=+\infty$) and IB-ERM ($\lambda=+\infty$) are
\begin{equation}\nonumber
\begin{aligned}
& w_1=0, w_2=0. 
\end{aligned}
\end{equation}

These calculation results are also consistent with the simulations in Appendix~\ref{appendix: A}.


\section{More causality analyses}
\label{appendix:causality others}
Given the theoretical setting in \Cref{sec:theory}, we have the following corollaries.

\textbf{Setting:} 
Consider several training environments $\etr=\{e_1,e_2,...\}$ and $\xx^e$ to be the observed input of $e\in \etr$.
We adopt an anti-causal framework \cite{DBLP:journals/corr/abs-1907-02893} with data generation process as follows:
\vspace{-1mm}
\begin{equation}\nonumber
\begin{aligned}
&\yy=\gamma^\top \hat\xx_{inv} +\eta_{y},\\
&\xx^{e}_{inv} = \hat\xx_{inv} + \eta^{e}_{inv}, \quad \xx^{e}_s = \hat\xx^{e}_s + \eta^{e}_s, \\ 
&\xx^e=S\left(\begin{smallmatrix}\xx^{e}_{inv} \\ \xx_s^{e} \end{smallmatrix}\right),
\vspace{-1mm}
\end{aligned}
\end{equation}
where $\gamma\in \RR^{d_{inv}}$ and $\gamma \ne \0$, the hidden invariant feature $\hat\xx_{inv}$ and the observed invariant feature $\xx^e_{inv}$ take values in $\RR^{d_{inv}}$, the hidden spurious feature $\hat\xx_s^{e}$ and the observed spurious feature $\xx_s^e$ take values in $\RR^{d_s}$, and $S: \RR^{(d_{inv}+d_s)}\to \RR^d$ is an inherent mapping to mix features.
The hidden spurious feature $\hat\xx_s^{e}$ is generated by $\yy$ with any \emph{non-invariant} relationship, $\eta^{e}_{inv}$ and $\eta^{e}_s$ are independent Gaussian with bounded mean and variance changed by environments, $\eta_{y}$ is an independent and invariant zero-mean Gaussian with bounded variance.   
As the directed acyclic graph (DAG) in \Cref{fig:3}(b) shows, the hidden invariant feature $\hat\xx_{inv}$ generates the true label $\yy$ and $\yy$ generates the hidden spurious feature $\hat\xx^{e}_s$.
In consideration of environmental noise, we can only observe the input $\xx^e$ which is a mixture of  $\xx^e_{inv}$ and $\xx^e_{s}$ after mapping. 
(Note that the observed feature is generated by applying environmental noise to the hidden feature.)
We follow the assumption from IRM~\cite{DBLP:journals/corr/abs-1907-02893}, i.e., assume that there exists a mapping $\tilde S: \RR^d \to \RR^{d_{inv}}$ such that $\tilde S (S (\begin{smallmatrix}\x_1 \\\x_2 \end{smallmatrix})) = \x_1$ for all $\x_1 \in \RR^{d_{inv}}, \x_2 \in \RR^{d_s}$.
and aim to learn a classifier to predict $\yy$ based on $\xx^e$, i.e., $f(\xx^e;\w)=h(g(\xx^e ; \Phi); \vv)$.

\begin{mycor}
\label{lem4}
If $\Phi$ elicits the desired invariant predictor $f(\cdot;\w)=\gamma^\top \tilde S (\cdot)$, there exist noisy environments $\{e_1,e_2\}$ such that 
\begin{equation}\nonumber
    \nabla_{\w}\R^{e_1}(\w)\ne \nabla_{\w}\R^{e_2}(\w).
\end{equation}
\end{mycor}
\begin{custompro}{C.1}
If $\Phi$ elicits the desired invariant predictor $f(\cdot;\w)=\gamma^\top \tilde S (\cdot)$ in noisy environments $\{e_1,e_2\}$, 
given square loss and the fixed ``dummy'' classifier $\vv=1$, we have 
\begin{equation}
\begin{aligned}
\frac{\partial \R^e(\w)}{\partial \vv|_{\vv=1}} &= \frac{\frac{1}{2}\EE_{\xx^e, \yy}[(f(\xx^e;\w)-\yy)^2]}{\partial \vv|_{\vv=1}}   \\
&=\frac{\frac{1}{2}\EE_{\xx^e, \yy}[(\vv|_{\vv=1} (\gamma^\top \hat\xx_{inv}+\gamma^\top\eta^e_{inv})-\gamma^\top\hat\xx_{inv}-\eta_y)^2]}{\partial \vv|_{\vv=1}} \\
&=\EE_{\xx^{e},y}\Big((\gamma^\top\hat\xx_{inv} + \gamma^\top \eta^e_{inv})(\gamma^\top\eta^e_{inv} - \eta_y) \Big) \\
&=\EE_{\xx^{e},y}\Big(\gamma^\top\eta_{inv}^e\gamma^\top\hat\xx_{inv} + (\gamma^\top \eta^e_{inv})^2 - \gamma^\top\hat\xx_{inv} \eta_{y} - \gamma^\top \eta^e_{inv}\eta_y \Big),
\end{aligned}
\end{equation}
where $e\in\{e_1,e_2\}$.

Obviously, when $\gamma\ne\0$, there exists $\eta_{inv}^{e_1}\ne \eta_{inv}^{e_2}$ such that $\frac{\partial \R^{e_1}(\w)}{\partial \vv|_{\vv=1}} \ne \frac{\partial \R^{e_2}(\w)}{\partial \vv|_{\vv=1}}$. 
\hfill $\square$
\end{custompro}
\Cref{lem4} shows that $||\nabla_{\w}\R^{e_1}(\w)-\nabla_{\w}\R^{e_2}(\w)||_2^2\to 0$ (IGA) may also be failed to find the optimal invariant predictor in noisy environments. 
Given different inherent losses, it seems unreasonable to enforce all gradients to be equal across environments.

\begin{mycor}
\label{lem5}
If $\Phi$ elicits the desired invariant predictor $f(\cdot;\w)=\gamma^\top \tilde S (\cdot)$, there exist noisy environments $\{e_1,e_2\}$ such that  
\begin{equation}\nonumber
    \Var(\nabla_{\w}\R(\xx^{e_1},\w))\ne \Var(\nabla_{\w}\R(\xx^{e_2},\w)).
\end{equation}
\end{mycor}
\begin{custompro}{C.2}
If $\Phi$ elicits the desired invariant predictor $f(\cdot;\w)=\gamma^\top \tilde S (\cdot)$ in noisy environments $\{e_1,e_2\}$, given square loss and the fixed ``dummy'' classifier $\vv=1$, we have 
\begin{equation}\nonumber
\begin{aligned}
\Var\left(\frac{\partial \R(\xx^e,\w)}{\partial \vv|_{\vv=1}}\right) =& \EE_{\xx^{e},y}\Big( \left[\gamma^\top\eta_{inv}^e\left( \gamma^\top\hat\xx_{inv}+\gamma^\top \eta^e_{inv}-\gamma_y \right)\right]^2 + \left(\gamma^\top \hat\xx_{inv}\eta_y\right)^2 \\
&- 2\gamma^\top \hat\xx_{inv}\eta_y\gamma^\top\eta_{inv}^e\left( \gamma^\top\hat\xx_{inv}+\gamma^\top \eta^e_{inv}-\gamma_y \right) \Big)  \\ 
&-\left[\EE_{\xx^{e},y}\Big(\gamma^\top\eta_{inv}^e\gamma^\top\hat\xx_{inv} + (\gamma^\top \eta^e_{inv})^2 - \gamma^\top\hat\xx_{inv} \eta_{y} - \gamma^\top \eta^e_{inv}\eta_y \Big)\right]^2,
\end{aligned}
\end{equation}
where $e\in\{e_1,e_2\}$.

Clearly, when $\gamma\ne\0$, there exists $\eta_{inv}^{e_1}\ne \eta_{inv}^{e_2}$ such that $\Var\left(\frac{\partial \R(\xx^{e_1},\w)}{\partial \vv|_{\vv=1}}\right) \ne \Var\left(\frac{\partial \R(\xx^{e_2},\w)}{\partial \vv|_{\vv=1}}\right)$. 
\hfill $\square$
\end{custompro}
Corollary~\ref{lem5} implies that looking for the optimal invariant predictor in noisy environments via $||\Var(\nabla_{\w}\R(\xx^{e_1},\w))-\Var(\nabla_{\w}\R(\xx^{e_2},\w))||_2^2\to 0$ (Fishr) may not always be successful,
for the reason that environmental inherent noises can affect the variance of gradients.

\begin{mycor}
\label{lem6}
Given $y\in\{-1,1\}$ and the fixed ``dummy'' classifier $\vv=1$, if $\Phi$ elicits the desired invariant predictor $f(\cdot;\w)=\gamma^\top \tilde S (\cdot)$, there exists $e$ in noisy environments such that 
\begin{equation}\nonumber
    \Var( g(\xx^e ; \Phi) |y)\ne 0.
\end{equation}
\end{mycor}
\begin{custompro}{C.3}
Given $y\in\{-1,1\}$ and the fixed ``dummy'' classifier $\vv=1$, if $\Phi$ elicits the desired invariant predictor $f(\cdot;\w)=\gamma^\top \tilde S (\cdot)$, we have
\begin{equation}\nonumber
    \Var(g(\xx^e ; \Phi)|y)=\Var\big((\gamma^\top\hat\xx_{inv}|y)+\gamma^\top \eta_{inv}^e\big).
\end{equation}
Obviously, we can find a $\eta_{inv}^e$ in noisy environments such that $\Var(g(\xx^e ; \Phi)|y)\ne 0$.
\hfill $\square$
\end{custompro}
Corollary~\ref{lem6} suggests that the IB penalty (IB-ERM) may also be unsuccessful to find the optimal invariant predictor in noisy environments.

\section{Proofs}
\label{appendix:proofs}

Here, we provide the proofs for Theorem~\ref{thm1}, Corollary~\ref{lem2} and Corollary~\ref{lem3}, respectively.

\begin{custompro}{3.1}
Assume that there exists a mapping $\tilde S: \RR^d \to \RR^{d_{inv}}$ such that $\tilde S (S (\begin{smallmatrix}\x_1 \\\x_2 \end{smallmatrix})) = \x_1$ for all $\x_1 \in \RR^{d_{inv}}, \x_2 \in \RR^{d_s}$.
Then, if $\Phi$ elicits the desired (optimal) invariant predictor $f(\cdot;\w)=\gamma^\top \tilde S (\cdot)$, we have
\begin{equation}
\begin{aligned}
\rho^e_{f,\yy}(\w)&=\EE_{\xx^e,y} [f(\xx^e;\w)y-\EE_{\xx^e}(f(\xx^e;\w))y]\\
&=\EE_{\xx^e, \yy}[\gamma^\top \tilde S(S(\begin{smallmatrix}\xx_{inv}^e \\\xx_s^{e} \end{smallmatrix})) \yy]-\EE[\gamma^\top \tilde S(S(\begin{smallmatrix}\xx_{inv}^e \\\xx_s^{e} \end{smallmatrix}))] \EE[\yy]\\
&=\EE_{\xx^e, \yy}[\gamma^\top \xx_{inv}^e \yy]-\EE[\gamma^\top \xx_{inv}^e] \EE[\yy]\\
&=\EE[(\gamma^\top \hat\xx_{inv})^2]-[\EE(\gamma^\top \hat\xx_{inv})]^2\\
&=\Var(\gamma^\top \hat\xx_{inv}),
\end{aligned}
\end{equation}
for all $e\in \eall$.
As $\Var(\gamma^\top \hat\xx_{inv})$ remains constant in all environments, we have $\Var(\rho^e_{f,\yy}(\w))=0$. 
\\Hence, proved.
\hfill $\square$
\end{custompro}

\begin{custompro}{3.2}
If $\Phi$ elicits the desired invariant predictor $f(\cdot;\w)=\gamma^\top \tilde S (\cdot)$, consider square loss and the fixed ``dummy'' classifier $\vv=1$, then
\begin{equation}
\begin{aligned}
\frac{\partial \R^e(\w)}{\partial \vv|_{\vv=1}} &= \frac{\frac{1}{2}\EE_{\xx^e, \yy}[(f(\xx^e;\w)-\yy)^2]}{\partial \vv|_{\vv=1}}   \\
&=\frac{\frac{1}{2}\EE_{\xx^e, \yy}[(\vv|_{\vv=1} (\gamma^\top \hat\xx_{inv}+\gamma^\top\eta^e_{inv})-\gamma^\top\hat\xx_{inv}-\eta_y)^2]}{\partial \vv|_{\vv=1}} \\
&=\EE_{\xx^{e},y}\Big((\gamma^\top\hat\xx_{inv} + \gamma^\top \eta^e_{inv})(\gamma^\top\eta^e_{inv} - \eta_y) \Big) \\
&=\EE_{\xx^{e},y}\Big(\gamma^\top\eta_{inv}^e\gamma^\top\hat\xx_{inv} + (\gamma^\top \eta^e_{inv})^2 - \gamma^\top\hat\xx_{inv} \eta_{y} - \gamma^\top \eta^e_{inv}\eta_y \Big).
\end{aligned}
\end{equation}
Obviously, when $\gamma\ne\0$, there exists $\eta_{inv}^{e}$ in noisy environments such that $\frac{\partial \R^{e}(\w)}{\partial \vv|_{\vv=1}} \ne \0$. 

$\quad$ \hfill $\square$
\end{custompro}

\begin{custompro}{3.3}
If $\Phi$ elicits the desired invariant predictor $f(\cdot;\w)=\gamma^\top \tilde S (\cdot)$, consider square loss, then
\begin{equation}
\begin{aligned}
\R^e(\w) &=  \frac{1}{2}\EE_{\xx^{e},y}\big[(\gamma^\top\hat\xx_{inv} + \gamma^\top \eta^e_{inv} - \gamma^\top\hat\xx_{inv} - \eta_y)^2 \big]\\
&=\frac{1}{2}\EE\big[(\gamma^\top \eta^e_{inv} - \eta_y)^2 \big],
\end{aligned}
\end{equation}
where $e\in\{e_1,e_2\}$.

Clearly, when $\gamma\ne\0$, there exists $\eta_{inv}^{e_1}\ne \eta_{inv}^{e_2}$ such that $\R^{e_1}(\w)\ne \R^{e_2}(\w)$. 
\hfill $\square$
\end{custompro}

\begin{figure}[t!]
\includegraphics[width=1
\textwidth]{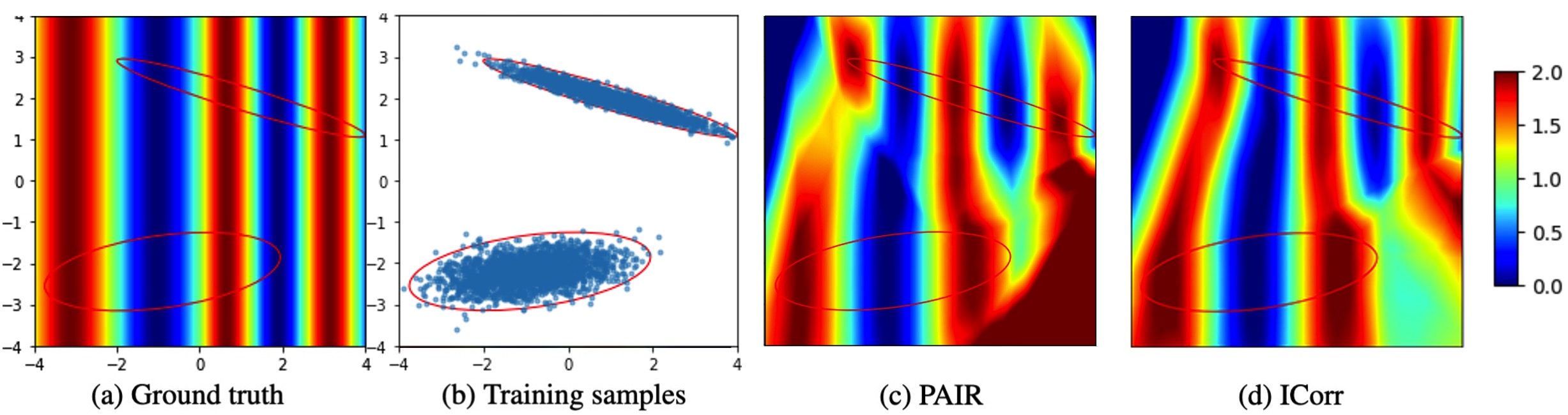}
\centering
\vspace{-6mm}
\caption{
Results of causal invariance~\cite{chen2023pareto} in noisy environments.
We run each method with 5 times and report the average losses: \textbf{(c)} PAIR 0.8164; \textbf{(d)} ICorr 0.6568.
}
\vspace{-4mm}
\label{fig:6}
\end{figure}

\section{More experiments and details}
\label{appendix:experiments}

\textbf{Experimental details:}
All experiments are implemented on NVIDIA A100 and AMD EPYC 7452 32-Core Processor.

For the experiment with ColoredMNIST, we use the exactly same setting as \url{https://github.com/capybaralet/REx_code_release/blob/master/Invari-antRiskMinimization/colored_mnist/main.py}, only replacing the IRMv1 penalty and VREx penalty with ICorr penalty and other penalties.

For the experiment with Circle Dataset, we use the exactly same setting as \url{https://github.com/hehaodele/CIDA/blob/master/toy-circle/main-half-circle.ipynb}, only applying noises to source domains and adding ICorr penalty term.


\textbf{Causal Invariance experiment:}
We then describe the definition of Causal Invariance specified by \cite{peters2016causal,DBLP:journals/corr/abs-1907-02893,pmlr-v130-kamath21a,chen2023pareto} as in Definition~\ref{def}.
\begin{mydef}
\label{def}
(Causal Invariance) Given a predictor $f(\cdot ; \w) = h(g(\cdot ; \Phi); \vv)$, the representation produced by the featurizer $\Phi$ is invariant over $\eall$ if and only if for all $e_1,e_2\in \eall$, it holds that
\begin{equation}
\EE_{\xx^{e_1},y}(y|g(\xx^{e_1};\Phi)=\zz)=\EE_{\xx^{e_2},y}(y|g(\xx^{e_2};\Phi)=\zz)
\end{equation}
for all $\zz\in \{ g(\xx^{e_1};\Phi)|e_1 \} \cap \{ g(\xx^{e_2};\Phi)|e_2 \}$.
\end{mydef}

Following \citet{chen2023pareto}, a regression example is designed with $\xx : \RR^2 \to y : \RR$. 
The input $\xx$ is with two dimensions, i.e., $\xx=(\xx_1,\xx_2)$, where $\xx_1$ represents horizontal axis and $\xx_2$ represents vertical axis in \Cref{fig:6}.
$\xx_1$ is designed to be the invariant feature and $\xx_2$ is designed to be the spurious feature.
Consider environmental inherent noises, we assume $y = sin(1.5*\xx_1) + 1$ for domains $\xx_1<0$ and $y = sin(2.5*\xx_1) + 1$ for domains $\xx_1\ge 0$.
All other settings are same with \url{https://github.com/LFhase/PAIR/blob/main/Extrapolation/pair_extrapolation.ipynb}.

We evaluate ICorr with Causal Invariance experiment from PAIR \cite{chen2023pareto}.
As shown in \Cref{fig:6}, $y = sin(1.5\cdot\xx_1) + 1$ for $\xx_1<0$ and $y = sin(2.5\cdot\xx_1) + 1$ for $\xx_1\ge 0$. 
$y$ is solely determined by $\xx_1$ (horizontal axis), while $\xx_2$ (vertical axis) does not influence the values of $y$.
Different colors represent different values of $y$. 
Note that we assume environmental noises influence domains $\xx_1<0$ with $sin(1.5\cdot\xx_1)$ and domains $\xx_1\ge 0$ with $sin(2.5\cdot\xx_1)$.
We sample two training areas as denoted by the ellipsoids colored in red (\Cref{fig:6}(b)).
With 5 repeats, ICorr achieves the lower average loss (0.6568) than PAIR (0.8164).

\textbf{Experiments with other noises:}
As shown in \Cref{tab:6666},  ICorr also gets a better performance in Poisson noisy and Uniform noisy environments.

\begin{table*}[t]
\centering
\caption{
Comparison of MLP on ColoredMNIST with varying training noises, i.e., first training environment without noise, second training environment with Poisson noise (with coefficient $0.1$) and Uniform noise ($[-0.1,0.1]$), respectively. 
We repeat each experiment with 20 times and report the best, worst and average accuracies (\%) on the test environment with Poisson noise and Uniform noise, respectively.
}
\label{tab:6666}
\vspace{0mm}
\scalebox{1}{
\begin{tabular}{cclcccccccc}
\specialrule{.1em}{.075em}{.075em} 
\multicolumn{1}{c}{\multirow{2}{*}{Test noise}} &&\multicolumn{1}{c}{\multirow{2}{*}{Method}} 
&& \multicolumn{3}{c}{$\{\0$, Poisson$\}_{\text{train}}$} && \multicolumn{3}{c}{$\{\0$, Uniform$\}_{\text{train}}$}        
\\
&&&& Best & Worst & Mean && Best & Worst & Mean         
\\
\cline{1-1} \cline{3-3} \cline{5-7} \cline{9-11}
\multirow{7}*{Poisson} && ERM && 49.55 & 10.13 & 26.58 && 49.64 & 9.73 & 20.49    \\
&& IRMv1 && 48.79 & 9.41 & 26.19 && 50.04 & 9.95 & 31.69    \\ 
&& VREx && 55.49 & 40.60 & 46.17 && 56.62 & 38.80 & 44.77  \\
&& CLOvE && 50.11 & 10.65 & 29.48 && 49.87 & 9.52 & 32.66       \\  
&& Fishr && 53.72 & 41.25 & 45.82 && 54.59 & 40.76 & 44.21       \\
&& ICorr && \textbf{60.95} & \textbf{44.18} & \textbf{53.31} && \textbf{59.13} & \textbf{47.73} & \textbf{53.01}   \\
\cline{1-1} \cline{3-3} \cline{5-7} \cline{9-11}
\multirow{7}*{Uniform} && ERM && 50.33 & 9.77 & 26.12 && 49.66 & 9.47 & 22.10     \\
&& IRMv1 && 49.91 & 9.46 & 26.23 && 49.48 & 9.69 & 29.17
   \\ 
&& VREx && 57.21 & 40.12 & 46.36 && 58.69 & 38.93 & 45.41
     \\
&& CLOvE && 51.80 & 10.33 & 30.18 && 50.12 & 9.70 & 32.47      \\  
&& Fishr && 53.98 & 40.93 & 46.79 && 54.66 & 41.27 & 46.10      \\ 
&& ICorr && \textbf{62.41} & \textbf{44.61} & \textbf{53.97} && \textbf{60.88} & \textbf{48.11} & \textbf{53.58}   \\
\specialrule{.1em}{.075em}{.075em}
\end{tabular}
}
\vspace{0mm}
\end{table*}

\end{appendices}


\bibliography{sn-bibliography}

\end{document}